\documentclass[lettersize]{IEEEtran}

\usepackage{amsmath,amsfonts}
\usepackage{algorithmic}
\usepackage{array}
\usepackage[caption=false,font=normalsize,labelfont=sf,textfont=sf]{subfig}
\usepackage{textcomp}
\usepackage{stfloats}
\usepackage{url}
\usepackage{verbatim}
\def\BibTeX{{\rm B\kern-.05em{\sc \kern-.025em b}\kern-.08em
    T\kern-.1667em\lower.7ex\hbox{E}\kern-.125emX}}
\usepackage{balance}

\usepackage{multirow}
\usepackage{tabularx,booktabs}
\usepackage[T1]{fontenc}
\usepackage{amssymb}% http://ctan.org/pkg/amssymb
\usepackage{pifont}% http://ctan.org/pkg/pifont
\newcommand{\cmark}{\ding{51}}%
\newcommand{\xmark}{\ding{55}}%
 \usepackage{graphicx}
 \usepackage{float}
\usepackage{graphbox}
\usepackage[normalem]{ulem}
\usepackage{arydshln}
\usepackage[dvipsnames]{xcolor}

\newif\ifcolorcomments
\colorcommentsfalse % <-- metti \colorcommentsfalse per disattivare i colori

\ifcolorcomments
  \expandafter\newcommand\csname rev1\endcsname[1]{\textcolor{magenta}{#1}}    
  \expandafter\newcommand\csname rev3\endcsname[1]{\textcolor{cyan}{#1}}     
\else
  \expandafter\newcommand\csname rev1\endcsname[1]{#1}
  \expandafter\newcommand\csname rev3\endcsname[1]{#1}
\fi

\begin{document}

%\title{Class-Agnostic Counting: A Survey on Advancements from Reference-Based to Open-World Text-Specified Approaches 
%for Counting Arbitrary Object Categories 
%in images
%}
\title{A Survey on Class-Agnostic Counting: Advancements from Reference-Based to Open-World Text-Guided Approaches}

\author{Luca Ciampi$^1$, Ali Azmoudeh$^{2}$, Elif Ecem Akbaba$^{2}$, Erdi Sarıtaş$^{2}$, Ziya Ata Yazıcı$^{2}$, \\ Hazım Kemal Ekenel$^{2,3}$, Giuseppe Amato$^1$, Fabrizio Falchi$^1$  \\
$^1$CNR-ISTI, Pisa, Italy \hfill $^2$Istanbul Technical University, Türkiye \hfill $^3$Division of Engineering, NYU Abu Dhabi, UAE
\thanks{This work was partially funded by: Spoke 8, Tuscany Health Ecosystem (THE) Project (CUP B83C22003930001), funded by the National Recovery and Resilience Plan (NRRP), within the NextGeneration Europe (NGEU) Program; Horizon Europe Research \& Innovation Programme under Grant agreement N. 101092612 (Social and hUman ceNtered XR - SUN project); PNRR - M4C2 - Investimento 1.3, Partenariato Esteso PE00000013 - "FAIR - Future Artificial Intelligence Research" - Spoke 1 "Human-centered AI", funded by European Union - NextGenerationEU.

%This work has been submitted to the IEEE for possible publication. Copyright may be transferred without notice, after which this version may no longer be accessible.
}}

%\markboth{Journal of \LaTeX\ Class Files,~Vol.~18, No.~9, September~2020}%
%{How to Use the IEEEtran \LaTeX \ Templates}

\maketitle

\IEEEpubidadjcol
\IEEEoverridecommandlockouts
\IEEEpubid{\makebox[\columnwidth]{Preprint. Accepted at Elsevier CVIU. \hfill} \hspace{\columnsep}\makebox[\columnwidth]{ }}

\begin{abstract}
Visual object counting has recently shifted towards class-agnostic counting (CAC), which addresses the challenge of counting objects across arbitrary categories---a crucial capability for flexible and generalizable counting systems. Unlike humans, who effortlessly identify and count objects from diverse categories without prior knowledge, most existing counting methods are restricted to enumerating instances of known classes, requiring extensive labeled datasets for training and struggling in open-vocabulary settings. In contrast, CAC aims to count objects belonging to classes never seen during training, operating in a few-shot setting.
In this paper, we present the first comprehensive review of CAC methodologies. We propose a taxonomy to categorize CAC approaches into three paradigms based on how target object classes can be specified: reference-based, reference-less, and open-world text-guided. Reference-based approaches achieve state-of-the-art performance by relying on exemplar-guided mechanisms. Reference-less methods eliminate exemplar dependency by leveraging inherent image patterns. Finally, open-world text-guided methods use vision-language models, enabling object class descriptions via textual prompts, offering a flexible and promising solution. Based on this taxonomy, we provide an overview of 30 CAC architectures and report their performance on gold-standard benchmarks, discussing key strengths and limitations. Specifically, we present results on the FSC-147 dataset, setting a leaderboard using gold-standard metrics, and on the CARPK dataset to assess generalization capabilities. Finally, we offer a critical discussion of persistent challenges, such as annotation dependency and generalization, alongside future directions. We believe this survey offers a valuable resource for researchers, capturing the evolution of CAC and providing insights to guide future developments in the field.
\end{abstract}


%%%%%%%%%%%%%%%%%%%%%%%%%%%%%%%%%%%%%%%%
% INTRODUCTION
%%%%%%%%%%%%%%%%%%%%%%%%%%%%%%%%%%%%%%%%
\section{Introduction}
\label{sec:introduction}

\begin{figure*}[!t]
    \centering
    \includegraphics[width=.8\linewidth]{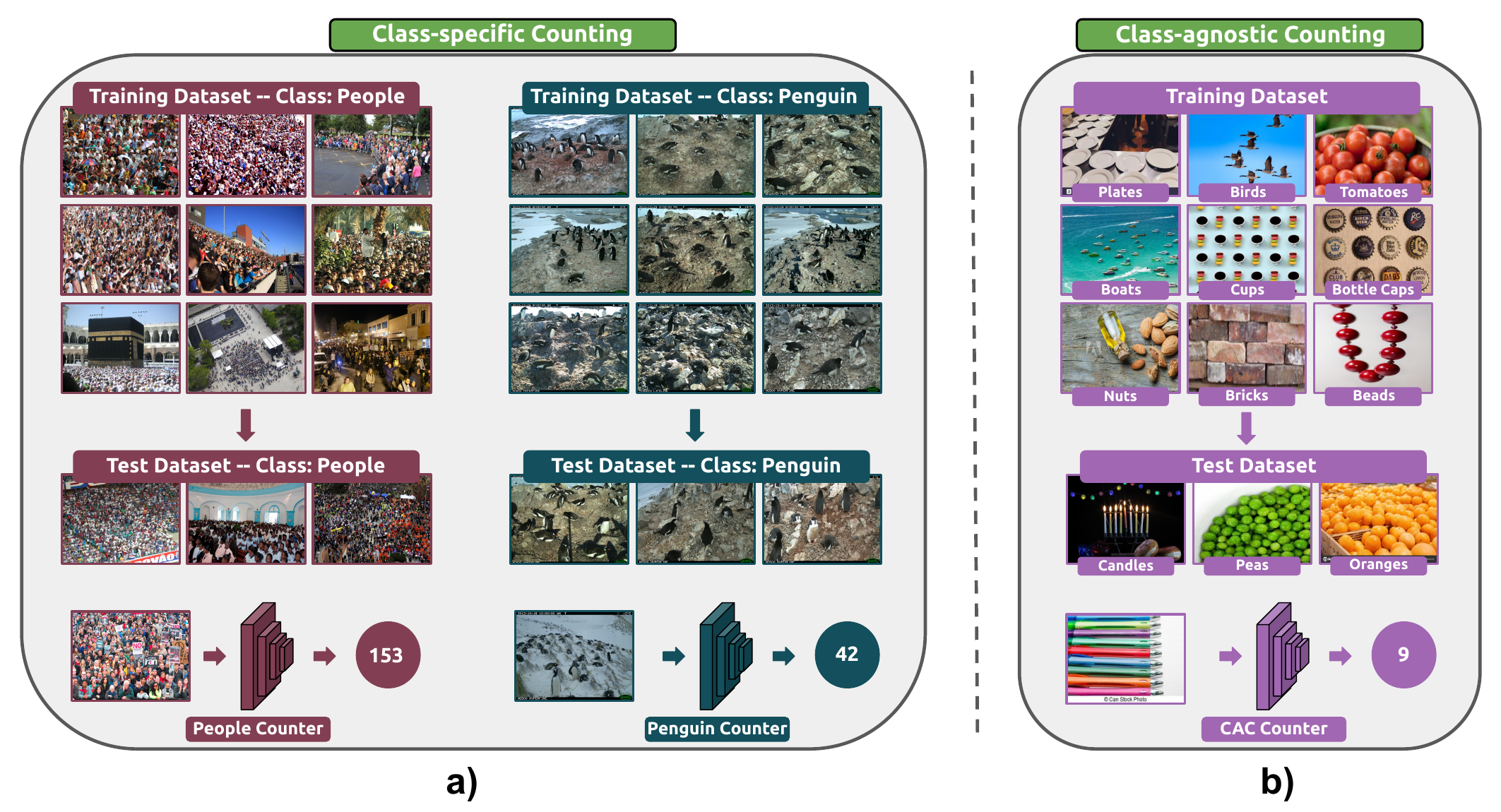}
    \caption{\csname rev1\endcsname{\textbf{Comparison between class-specific and class-agnostic counting.} On the left, we show examples of class-specific counting, where each object category (e.g., cars, people, animals) requires a dedicated model trained on large, annotated datasets. These models are tailored to recognize and count only the specific class they were trained on. On the right, we present the class-agnostic counting approach, which aims to estimate the number of objects regardless of their semantic category. In this setting, the counting model is trained on a set of object classes and evaluated on entirely different, unseen classes, demonstrating its ability to generalize across categories. This paradigm significantly reduces the need for exhaustive per-class annotations and enables broader applicability in real-world scenarios.}}
    \label{fig:teaser}
\end{figure*}

The goal of object counting is to automatically estimate the number of object instances in still images or video frames~\cite{DBLP:conf/nips/LempitskyZ10}. This challenging task has emerged as a prominent research focus within the realm of computer vision, owing to its vast array of practical applications and interdisciplinary significance. \csname rev1\endcsname{For instance, in crowd analysis~\cite{DBLP:conf/cvpr/SamSB17,DBLP:conf/mm/BoominathanKB16,DBLP:conf/cvpr/ZhangLWY15,DBLP:conf/cvpr/ShiZLCYCZ18,DBLP:conf/cvpr/LiuSF19,DBLP:conf/wacv/HossainHCW19,DBLP:conf/eccv/CaoWZS18,DBLP:conf/iscc/AvvenutiBCFGM22,DBLP:journals/eswa/BenedettoCCFGA22,DBLP:journals/tip/ZhaoL23}, it supports event monitoring with implications across social~\cite{DelRe2013}, psychological~\cite{ca55bf97-727f-3539-95ea-88cef713b2cc}, political~\cite{doi:10.1177/0049124116629166}, and security domains~\cite{DBLP:journals/eor/AbdelghanyAMA14}. In agriculture, it enables livestock and pest monitoring~\cite{DBLP:journals/cea/TianGCWLM19,DBLP:conf/avss/SarwarGPPL18,DBLP:journals/ecoi/CiampiZICBFAC23}, and plant emergence assessment~\cite{DBLP:conf/icaisc/AlkhudaydiZl19,DBLP:journals/corr/LuC0ZS17,DBLP:conf/iccvw/PoundAWPF17}. Other applications include traffic analysis~\cite{DBLP:journals/access/DaiSWFYZL19,DBLP:journals/access/TayaraSC18,DBLP:conf/iccv/ZhangWCM17,DBLP:conf/iscc/AmatoCFG19,DBLP:journals/eswa/CiampiGCFVA22}, inventory control~\cite{DBLP:conf/ist/BalaskaFKG22}, and biomedical imaging~\cite{DBLP:conf/iccvw/CohenBGLB17,DBLP:journals/mia/CiampiCTMLSAPG22,DBLP:conf/bibe/ZhuSQH17,DBLP:journals/cmbbeiv/XieNZ18}.} 
\csname rev1\endcsname{However, unlike humans---who can naturally identify what is worth counting and recognize repetitions even among unfamiliar objects---most AI-based methods are constrained to predefined categories. These approaches typically rely on prior knowledge of the object class and require dedicated models trained on large annotated datasets~\cite{DBLP:conf/cvpr/RanjanSNH21} (see also the left part of Fig.~\ref{fig:teaser}). Even recent general-purpose agents and MLLMs, such as LLaVA~\cite{DBLP:conf/nips/LiuLWL23a}, MiniGPT4~\cite{DBLP:journals/corr/abs-2304-10592}, and mPLUG-owl~\cite{DBLP:journals/corr/abs-2304-14178}, struggle with counting objects from open-vocabulary categories~\cite{DBLP:conf/nips/YinWCSLLH0S0SO23}.}

Very recently, research about object counting has shifted towards a new trend that aims to consider arbitrary object categories and attempts to alleviate the problem of the annotation burden (see right part of Fig.~\ref{fig:teaser}). The first attempt of creating a model able to count any class, i.e., to perform Class-Agnostic Counting (CAC), \csname rev1\endcsname{was proposed by Lu and Zisserman through the Generic Matching Network (GMN)~\cite{DBLP:conf/accv/LuXZ18}, which formulates counting as a matching task by exploiting image self-similarity---since objects to be counted often repeat within the same image. GMN is pre-trained on tracking data and includes an adaptation module for domain transfer. However, this module requires dozens to hundreds of examples and performs poorly on novel classes without adaptation. The first work to significantly depart from standard class-specific counting was introduced by Ranjan et al.~\cite{DBLP:conf/cvpr/RanjanSNH21}, who introduced a few-shot regression approach to counting. In this setting, the inputs are an image and a set of exemplars---bounding boxes around objects of interest from the same image---used as object prototypes. At inference time, similar to the few-shot classification paradigm~\cite{DBLP:conf/icml/FinnAL17,DBLP:conf/nips/VinyalsBLKW16}, the objects to be counted belong to unseen categories, making few-shot counting fundamentally different from traditional counting, where training and test classes match. Notably, this paradigm does not learn from limited data via domain adaptation~\cite{DBLP:conf/cvpr/WangGL019,DBLP:conf/visapp/CiampiSCGA21} or uncertainty modeling~\cite{DBLP:conf/aaai/OhOR20}, but instead uses supervised data to train a class-agnostic model applicable to novel categories---making it data-efficient for unseen classes.}

%The typical formulation involves training an object counting network using images and density maps as the training targets, eventually together with so-called exemplars~\cite{DBLP:conf/cvpr/RanjanSNH21}; the latter is defined as additional images containing single examples of the objects to be counted and are used by the network as a prototype for detecting that objects in the whole images. At test time, the classes of the objects to be counted are entirely different from the ones seen during training, making this task very different from the typical counting task, where the training and test classes are the same. It is worth noting that these methods don't explicitly learn from limited data. Instead, this strategy involves using a lot of fully supervised data to learn a class-agnostic counting model that can then later be applied to unseen object categories with limited additional data required. In this way, it is a limited data method with respect to the unseen object categories.

\begin{figure*}[!t]
    \centering
    \includegraphics[width=.9\linewidth]{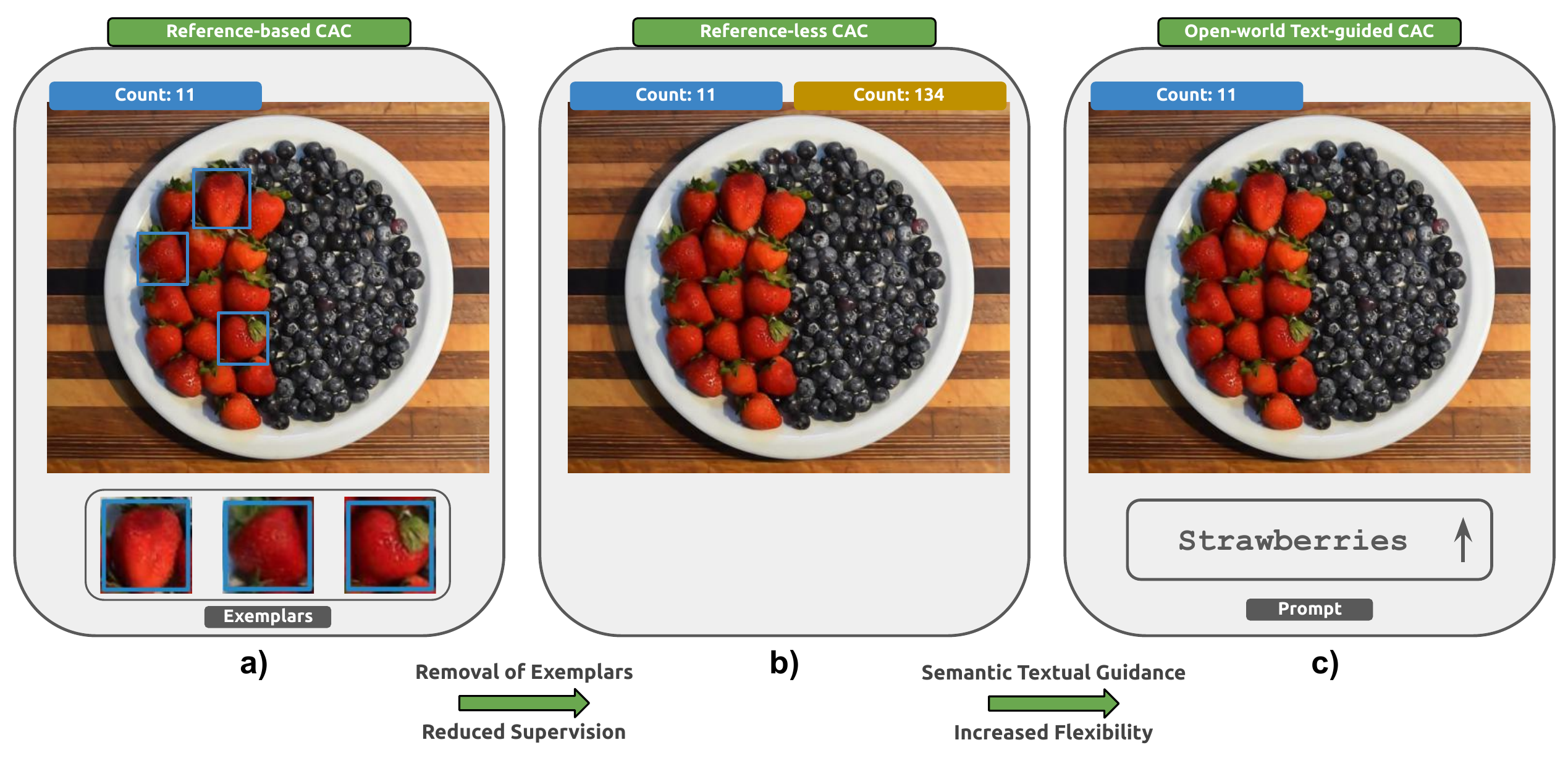}
    \caption{\textbf{Overview of class-agnostic counting paradigms.} We propose a taxonomy to classify existing CAC methodologies: a) \textit{Reference-based} approaches rely on annotated bounding box exemplars, which serve as visual prototypes for the object classes to be counted; b) \textit{Reference-less} techniques relax these requirements, enabling models to automatically identify the dominant class(es) to be counted; and c) \textit{Open-world Text-guided} methodologies allow the use of textual descriptions to specify the object classes to be considered. \csname rev3round2\endcsname{The arrows indicate the conceptual evolution of CAC paradigms, highlighting the progressive reduction of human supervision and the increasing flexibility and semantic control.}}
    \label{fig:class-agnostic-paradigms}
\end{figure*}

For the first time, this survey explores and analyzes the leading works existing in the literature tackling class-agnostic counting. \csname rev1\endcsname{Despite being a recent task, it has already attracted attention from top-tier venues, yielding progressively better results.} Furthermore, its formulation has slightly changed and evolved compared to its original version, gradually reducing human intervention in the overall process, i.e., the need for annotated exemplars, either at training or inference time.
%. Progressively, CAC reduced the initial human intervention in the overall process, i.e., the need for annotated exemplars, either at training or inference time. 
In this context, our contributions are fourfold: (i) we propose a taxonomy to categorize CAC methods, bringing clarity to the literature; (ii) we provide an overview of the architectures of 30 CAC approaches; (iii) we present and compare the performance of the examined techniques on gold-standard benchmarks; (iv) we critically discuss the results from multiple perspectives.
%, highlighting the strengths and limitations of the considered methodologies, the persistent challenges in the field, and potential directions for future research.

Specifically, the proposed taxonomy organizes the surveyed CAC methods into three categories, also illustrated in Fig.~\ref{fig:class-agnostic-paradigms}.
The first category follows the previously mentioned \textit{reference-based} or \textit{few-shot} paradigm~\cite{DBLP:conf/cvpr/RanjanSNH21}, where annotated bounding box exemplars are required during both training and inference.
Subsequently, CAC evolved into a formulation that shapes our second category, where manually specifying object categories is no longer necessary; the algorithm automatically identifies the dominant class(es) to be counted. This approach is known in the literature as \textit{reference-less} or \textit{exemplar-free}. Finally, the most recent paradigm, referred to as \textit{open-world text-guided}, \textit{prompt-based}, or \textit{zero-shot}, allows the system to take natural language descriptions of object classes as input, going hand in hand with the latest vision-language foundation models, and constitutes our third category. 
\csname rev3round2\endcsname{These three paradigms reflect the evolution of CAC methodologies toward reduced human supervision and increased flexibility. Reference-based methods, while effective, are not well-suited for autonomous systems, as they require users to manually provide exemplar inputs for each new category at inference time. Reference-less approaches address this limitation by removing the need for exemplars altogether, inferring the dominant object class directly from image patterns. However, they still lack semantic control and cannot incorporate user intent, making them suitable only when the target class aligns with the most visually repetitive category in the scene. Text-guided methods extend this trajectory by introducing a more natural and flexible way to specify the target category---closer to how humans describe objects---while leveraging the semantic capabilities of textual prompts. Rather than forming isolated categories, each paradigm represents a conceptual step forward, progressively addressing the limitations of the previous one and contributing to a more flexible and interaction-aware framework for CAC.}

%Based on this taxonomy, we present an overview of the architectures of the examined methods and show their performance on established gold-standard benchmarks from the literature. In particular, we consider the FSC-147 dataset~\cite{DBLP:conf/cvpr/RanjanSNH21}, which includes over 6,000 images across 147 object categories, and the CARPK dataset~\cite{DBLP:conf/iccv/HsiehLH17}, explicitly designed for vehicle counting and used to assess cross-dataset generalization. Lastly, we critically discuss the results, highlighting the strengths and limitations of the considered methodologies, the persistent challenges in the field, and potential directions for future research.

Based on this taxonomy, we present an overview of the architectures of the examined methods and show their performance on established benchmarks from the literature. \csname rev4round3\endcsname{In particular, we consider the FSC-147 dataset~\cite{DBLP:conf/cvpr/RanjanSNH21}, introduced together with the few-shot CAC formulation and now regarded as the gold-standard benchmark for this task.} FSC-147 includes over 6,000 images across 147 object categories. We also consider the CARPK dataset~\cite{DBLP:conf/iccv/HsiehLH17}, explicitly designed for vehicle counting and used to assess cross-dataset generalization. Lastly, we critically discuss the results, highlighting the strengths and limitations of the considered methodologies, the persistent challenges in the field, and potential directions for future research.

%and opening the door for accounting also images having multiple dominant categories.
%Starting from the above-mentioned \textit{reference-based} paradigm~\cite{DBLP:conf/cvpr/RanjanSNH21}, where the annotated bounding box exemplars have been gradually decreased from three to one, class-agnostic counting switched to a \textit{reference-less} or \textit{zero-shot} formulation. 
%In this latter case, there is no more need to specify the object categories to be counted manually; the algorithm is responsible for automatically considering the most dominant class(es). The subsequent paradigm, named \textit{prompt-based} class-agnostic counting, allowed the counting system to take natural language descriptions of the object class as input, going hand in hand with the latest vision-language foundation models and opening the door for accounting also images having multiple dominant categories.
%tried to overcome the limitations of reference-less setting, enabling the counting system to receive as input an arbitrarily user-specified class name in the form of text, thus opening the door for accounting also images having multiple dominant categories. 

We organize the rest of this work as follows. In Sec.~\ref{sec:existing-surveys-scope}, we review the most influential surveys related to object counting, emphasizing the scope of the current work and outlining the search strategy. Sec.~\ref{sec:existing_works} presents the existing methods according to the taxonomy introduced earlier. Next, in Sec.~\ref{sec:dataset}, we provide an overview of the available datasets. In Sec.~\ref{sec:results}, we present the results along with a critical discussion. Finally, in Sec.~\ref{sec:conclusion}, we conclude the paper with insights into potential directions for future research.

%%%%%%%%%%%%%%%%%%%%%%%%%%%%%%%%%%%%%%%%
% EXISTING SURVEYS
%%%%%%%%%%%%%%%%%%%%%%%%%%%%%%%%%%%%%%%%
%\section{Existing Surveys and Scope}
\section{Existing Surveys, Scope, and Search Strategy}
\label{sec:existing-surveys-scope}

%\subsection{Existing Surveys}
\noindent While there are several survey papers reviewing counting methods present in the computer vision literature, most of them concentrate on specific object categories, with a particular emphasis on crowd counting. The most cited survey is~\cite{DBLP:journals/prl/SindagiP18}. In this work, Sindagi et al. revised the papers up to 2017 in detail, focusing on crowd counting and considering both traditional methodologies and techniques based on convolutional neural networks. Furthermore, they described the most used crowd counting datasets, presenting and discussing the results obtained against these benchmarks. More recently, Li et al. published a similar, more updated survey in~\cite{DBLP:journals/paa/LiHZLL21}, retracing the last 20 years of object counting, mainly focusing on crowd counting with density estimation using convolutional neural networks. A plethora of other surveys or short surveys have been published over the last few years: a non-exhaustive list includes~\cite{DBLP:journals/corr/abs-2003-12783,DBLP:journals/bdcc/GouiaaAS21,PATWAL20232448,DBLP:journals/ijon/BaiMC22,DBLP:journals/ivc/KhanMH23,10.1007/978-981-15-4409-5_76,DBLP:journals/sensors/HassenMT22,DBLP:journals/mta/YangZ23,DBLP:journals/itc/HaoDMLF23,DBLP:journals/ijon/FanZZLZW22}. \csname rev4round3\endcsname{Even if the authors of these works look at the topic from different angles, considering various taxonomies and perspectives, they consistently focus on single-class object counting.}
%, especially crowd counting.

\csname rev1round2\endcsname{In contrast, our survey is the first to provide a comprehensive and structured overview of CAC methodologies. To the best of our knowledge, the only other survey that briefly touches on CAC is~\cite{D_Alessandro_2023}, which covers only a few methods---likely due to the early stage of the task at the time of writing---and does not offer a systematic taxonomy or performance analysis. Conversely, our survey introduces several distinct and novel contributions: it organizes CAC methods according to an original taxonomy, systematically analyzes state-of-the-art approaches, compares their performance across benchmarks, and discusses their limitations and open challenges.}

%To select papers for this survey, we began by identifying the gold-standard benchmarks for the CAC task. We then conducted a comprehensive search on Google Scholar for papers that cited these benchmarks. To ensure the quality and reliability of the sources, we included only peer-reviewed published papers in our selection, \csname rev1\endcsname{excluding non-peer-reviewed preprints from arXiv unless they provided freely available implementations to ensure reproducibility.}

\csname rev4round3\endcsname{To select papers for this survey, we conducted a comprehensive search on Google Scholar for works citing FSC-147, the gold-standard benchmark introduced together with the few-shot CAC formulation by Ranjan et al.~\cite{DBLP:conf/cvpr/RanjanSNH21} (as mentioned in Sec.~\ref{sec:introduction}). The search covered publications up to the end of 2024; the survey was completed in January 2025 before submission. Only peer-reviewed published papers were considered, excluding non-peer-reviewed preprints from arXiv unless they provided freely available implementations to ensure reproducibility.}

\begin{figure*}[!t]
    \centering
    \includegraphics[width=.99\linewidth]{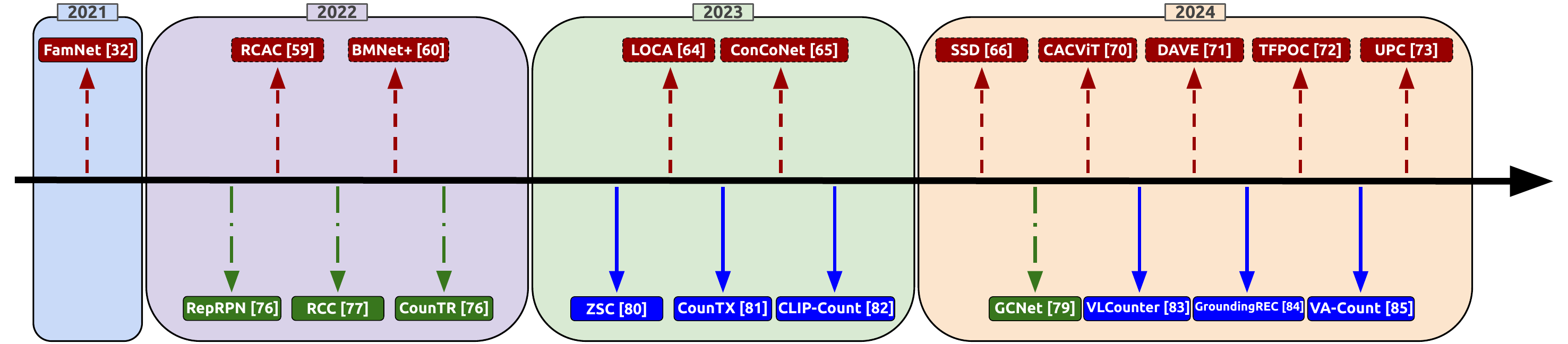}
    \caption{\csname rev1\endcsname{\textbf{Overview of selected CAC methodologies over time.}We present a chronological overview of representative Class-Agnostic Counting (CAC) methods, highlighting milestone contributions from 2021 to 2024. \csname rev4round3\endcsname{Each method is encoded using both color and arrow style according} to the taxonomy introduced in Fig.~2: \colorbox{BrickRed}{red} \csname rev4round3\endcsname{dashed arrows} for \textit{reference-based} approaches that rely on annotated exemplars; \colorbox{OliveGreen}{green} \csname rev4round3\endcsname{dash-dot arrows} for \textit{reference-less} methods that infer dominant object classes without manual input; and \colorbox{Blue}{blue} \csname rev4round3\endcsname{solid arrows} for \textit{open-world text-guided} approaches that use natural language prompts to specify target classes. This timeline illustrates the evolution of CAC paradigms over time, showing the shift from exemplar-dependent models to more flexible and generalizable solutions.}}
    \label{fig:temporal_line}
\end{figure*}

\section{Existing Approaches}
\label{sec:existing_works}
\csname rev1\endcsname{In the following, we review existing methodologies by grouping them into three categories based on our proposed taxonomy: reference-based, reference-less, and open-world text-guided approaches. While some methods include additional modules or architectural tweaks that may span multiple categories, each is classified under its primary type, with cross-category features discussed where relevant.
Table~\ref{tab:existing_methods} summarizes the surveyed techniques, while Fig.~\ref{fig:temporal_line} highlights key milestones over time.}

\subsection{Problem Formulation}
\label{sec:sec:problem-formulation}

\noindent \csname rev1\endcsname{State-of-the-art class-specific counting relies on supervised deep learning, typically following two strategies: detection-based and regression-based. Detection approaches identify individual instances using object detectors~\cite{DBLP:conf/iscc/AmatoCFG19,DBLP:journals/eswa/CiampiGCFVA22,DBLP:conf/avss/SarwarGPPL18,DBLP:journals/eswa/BenedettoCCFGA22}, while regression methods estimate object counts by learning a mapping from image features to a density map~\cite{DBLP:conf/nips/LempitskyZ10,DBLP:journals/corr/LuC0ZS17,DBLP:conf/eccv/ArtetaLZ16,Padubidri2021}, whose pixel values are integrated to obtain the final count.}
Detectors, while conceptually simpler and capable of handling a wide range of object classes due to the abundance of suitable annotated datasets, struggle with occluded objects. This limitation is the primary reason density-based approaches have largely replaced detection-based techniques. Thus, also most class-agnostic counting techniques follow the density-based paradigm, and CAC existing benchmarks reflect this in terms of label availability. Precisely, the gold standard labels needed for the supervised training of density regressors are dots (points), i.e., ground truth corresponds to binary spatial images with 1's at the center location of the objects, indicating their existence, and 0's at other locations without the objects. \csname rev1\endcsname{Since directly optimizing a loss on these sparse labels is challenging, most methods generate smoothed density maps by placing Gaussian kernels $G_\sigma$ at each dot location, using a fixed or adaptive $\sigma$ based on object size~\cite{DBLP:conf/cvpr/ZhangZCGM16,DBLP:conf/nips/LempitskyZ10}.}
%It is worth noting that approaches relying on detectors undergo severe limitations in images presenting several occluded objects, such as in typical scenarios characterizing the object counting task, and this represents the main reason why density-based approaches have supplanted detection-based techniques. 
%The same work that defined the density-based counting formulation~\cite{DBLP:conf/nips/LempitskyZ10} also proposed dots (points) as the gold standard labels needed for the supervised training of density regressors. Precisely, ground truth corresponds to binary spatial images with 1's at the objects' center location, indicating their existence, and 0's at other locations without the objects. Since it is difficult to train a density regressor with a loss directly based on dot annotations, most existing works generate smoothed target density maps by superimposing Gaussian kernels $G_\sigma$ centered in these dot-annotated locations and using a fixed or adaptive spread parameter $\sigma$, depending on the typical object size in the dataset~\cite{DBLP:conf/cvpr/ZhangZCGM16,DBLP:conf/nips/LempitskyZ10}. 
%However, even if dotting emulates the natural human technique of counting objects, producing thousands of ground truth samples entails a tremendous human effort and, consequently, puts a damper on the large-scale applicability of current counting solutions involving new scenarios and new object categories to be considered. 

\begin{table*}[!ht]
    \caption{\textbf{List of the considered methods.} We summarize the methodologies discussed in this survey, along with some relevant associated details. The methods are categorized based on our proposed taxonomy---reference-based, reference-less, and open-world text-guided \\}
    \label{tab:existing_methods}
    \setlength{\tabcolsep}{2pt}
    \small
    \newcolumntype{C}{>{\centering\arraybackslash}m{.1\linewidth}}
    \newcolumntype{Z}{>{\centering\arraybackslash}m{.22\linewidth}}
    \centering\begin{tabularx}{0.98\linewidth}{lZCCCCC}
            \toprule
            Method & Venue & Year & Code & \csname rev3\endcsname{Pretrained Model} & Point-level Supervision & Test-time Adaptation \\
            \midrule
            \midrule
            \textit{Reference-based} & & & & & & \\
            FamNet~\cite{DBLP:conf/cvpr/RanjanSNH21} & CVPR & 2021 & \cmark & \textcolor{black}{\cmark} & \cmark & \cmark \\
            CFOCNet~\cite{DBLP:conf/wacv/YangSHC21} & WACV & 2021 & \xmark & \textcolor{black}{\xmark} & \cmark & \xmark \\
            VCN~\cite{DBLP:conf/cvpr/RanjanH22} & CVPRW & 2022 & \xmark & \textcolor{black}{\xmark} & \cmark & \cmark \\
            RCAC~\cite{DBLP:conf/eccv/GongZ0DS22} & ECCV & 2022 & \cmark & \textcolor{black}{\xmark} & \cmark & \xmark \\
            BMNet+~\cite{DBLP:conf/cvpr/Shi0FL022} & CVPR & 2022 & \cmark & \textcolor{black}{\cmark} & \cmark & \xmark \\
            SPDNC~\cite{DBLP:conf/bmvc/LinYM0LLHYC22} & BMVC & 2022 & \cmark & \textcolor{black}{\cmark} & \cmark & \xmark \\
            MACnet~\cite{DBLP:journals/prl/McCarthyVOTAH23} & Pattern Recognit. Lett. & 2023 & \xmark & \textcolor{black}{\xmark} & \cmark & \cmark \\
            SAFECount~\cite{DBLP:conf/wacv/YouYLLCL23} & WACV & 2023 & \cmark & \textcolor{black}{\cmark} & \cmark & \xmark \\
            LOCA~\cite{DBLP:conf/iccv/EukicLZK23} & ICCV & 2023 & \cmark & \textcolor{black}{\cmark} & \cmark & \xmark \\
            ConCoNet~\cite{DBLP:journals/prl/SolivenVOTAH23} & Pattern Recognit. Lett. & 2023 & \xmark & \textcolor{black}{\xmark} & \cmark & \cmark \\
            %ASFNet~\cite{10145320} & - & - & - & - \\
            %SAM~\cite{DBLP:journals/corr/abs-2304-10817} & arXiv & 2023 & \cmark & \xmark \\
            SSD~\cite{DBLP:conf/ijcai/Xu0Z24} & IJCAI & 2024 & \cmark & \textcolor{black}{\xmark} & \cmark & \xmark \\
            SATCount~\cite{DBLP:journals/nn/WangYWLC24} & Neural Networks & 2024 & \xmark & \textcolor{black}{\xmark} & \cmark & \xmark \\
            CSTrans~\cite{DBLP:journals/pr/GaoH24} & Pattern Recognition & 2024 & \cmark & \textcolor{black}{\cmark} & \cmark & \xmark \\
            CountVers~\cite{DBLP:journals/kbs/YangCDWZ24} & Knowledge-Based Systems & 2024 & \xmark & \textcolor{black}{\xmark} & \cmark & \xmark \\ 
            CACViT~\cite{DBLP:conf/aaai/WangX0024} & AAAI & 2024 & \cmark & \textcolor{black}{\xmark} & \cmark & \xmark \\
            DAVE~\cite{DBLP:conf/cvpr/PelhanLZK24} & CVPR & 2024 & \cmark & \textcolor{black}{\cmark} & \cmark & \xmark \\
            TFPOC~\cite{DBLP:conf/wacv/Shi0Z24} & WACV & 2024 & \cmark & \textcolor{black}{\cmark} & \xmark & \xmark \\
            UPC~\cite{DBLP:conf/aaai/0018C24} & AAAI & 2024 & \xmark & \textcolor{black}{\xmark} & \cmark & \xmark \\
            PseCo~\cite{DBLP:conf/cvpr/HuangD0ZS24} & CVPR & 2024 & \cmark & \textcolor{black}{\cmark} & \cmark & \xmark \\
            \textcolor{black}{CountDiff~\cite{DBLP:conf/eccv/HuiWRL24}} & \textcolor{black}{ECCV} & \textcolor{black}{2024} & \textcolor{black}{\xmark} & \textcolor{black}{\xmark} & \textcolor{black}{\cmark} & \textcolor{black}{\cmark} \\
            %\textcolor{darkgreen}{GeCo~\cite{DBLP:conf/nips/PelhanLZK24}} & \textcolor{darkgreen}{NeurIPS} & \textcolor{darkgreen}{2024} & \textcolor{darkgreen}{\cmark} & \textcolor{darkgreen}{\cmark} & \textcolor{darkgreen}{\cmark} & \textcolor{darkgreen}{\xmark} \\ 
            %CFENet~\cite{DBLP:journals/ivc/ZhangZCWH25} & Image and Vision Computing & 2025 & \cmark & \xmark & \xmark \\
            % CAP~\cite{10671540} & ICSIP & 2024 & \xmark & \cmark & \xmark \\
            %MEAMNet~\cite{DBLP:journals/prl/ZhangHZLCL24} & Pattern Recognit. Lett. & 2024 & \xmark & \cmark & \xmark \\ 
            % CountGD~\cite{DBLP:journals/corr/abs-2407-04619} & arXiv & 2024 & \cmark & \cmark & \xmark \\ 
            \midrule
            \textit{Reference-less} & & & & & & \\
            RepRPN-Counter~\cite{DBLP:conf/accv/RanjanN22} & ACCV & 2022 & \cmark & \textcolor{black}{\xmark} & \cmark & N/A \\
            RCC~\cite{DBLP:journals/corr/abs-2205-10203} & arXiv & 2022 & \cmark & \textcolor{black}{\xmark} & \xmark & N/A \\
            CounTR~\cite{DBLP:conf/bmvc/LiuZZX22} & BMVC & 2022 & \cmark & \textcolor{black}{\xmark} & \cmark & N/A \\
            GCNet~\cite{DBLP:journals/pr/WangLZTG24} & Pattern Recognition & 2024 & \xmark & \textcolor{black}{\xmark} & \xmark & N/A \\
            \midrule
            \textit{Open-world Text-guided} & & & & & & \\
            ZSC~\cite{DBLP:conf/cvpr/XuL0RS23} & CVPR & 2023 & \cmark & \textcolor{black}{\xmark} & \cmark & \xmark \\
            CounTX~\cite{AminiNaieni23} & BMVC & 2023 & \cmark & \textcolor{black}{\cmark} & \cmark & \xmark \\
            CLIP-Count~\cite{DBLP:conf/mm/JiangLC23} & ACM MM & 2023 & \cmark & \textcolor{black}{\cmark} & \cmark & \xmark \\
            VLCounter~\cite{DBLP:conf/aaai/KangMKH24} & AAAI & 2024 & \cmark & \textcolor{black}{\cmark} & \cmark & \xmark \\
            GroundingREC~\cite{DBLP:conf/cvpr/DaiLC24} & CVPR & 2024 & \cmark & \textcolor{black}{\cmark} & \cmark & \xmark \\
            VA-Count~\cite{DBLP:conf/eccv/ZhuYYGWZH24} & ECCV & 2024 & \cmark & \textcolor{black}{\xmark} & \cmark & \xmark \\
            %\textcolor{darkgreen}{CountGD~\cite{DBLP:conf/nips/Amini-NaieniHZ24}} & \textcolor{darkgreen}{NeurIPS} & \textcolor{darkgreen}{2024} & \textcolor{darkgreen}{\cmark} & \textcolor{darkgreen}{\cmark} & \textcolor{darkgreen}{\cmark} & \textcolor{darkgreen}{\xmark} \\ 
            %T-Rex2~\cite{DBLP:journals/corr/abs-2403-14610} & arXiv & 2024 & \cmark & \cmark & - \\
            \bottomrule \\
    \end{tabularx} 
\end{table*}

\begin{figure*}[!t]
    \centering
    \includegraphics[width=.98\linewidth]{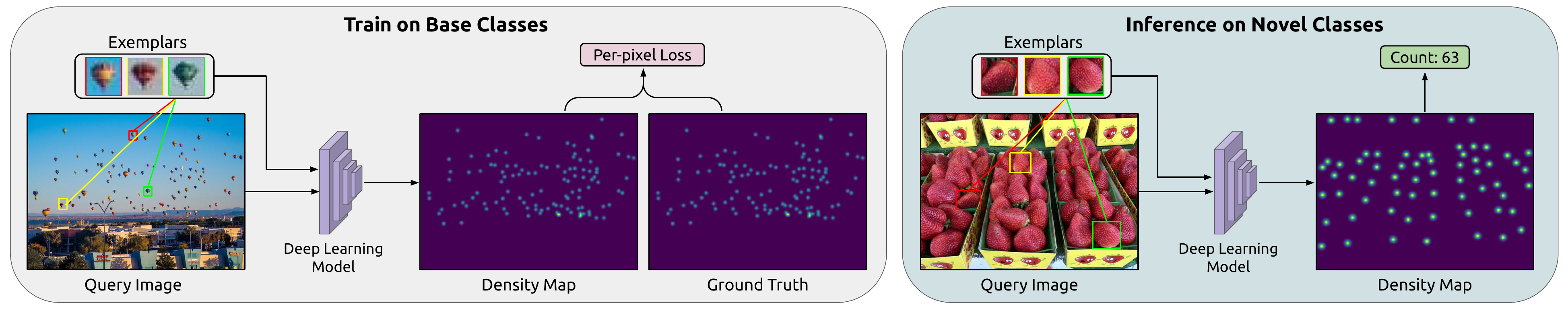}
    \caption{\textbf{Train and inference high-level overview of reference-based CAC approaches relying on density regression.} Following the notation of this survey, CAC models are fed with query images $I_i$ and $K$ exemplars expressed as bounding box coordinates $B^E = \{b_i\}_{i=1:k} \in \mathbb{R}^{4}$. Usually, $K = 3$. Exemplars belong to several object classes, which are different at training and test time. The model is in charge of (i) computing feature representations from these inputs that should be agnostic to specific object classes, and (ii) predicting density maps $D^{map}$ from these feature representations. Density-based methods are the standard approach for object counting in crowded scenes. Training typically involves minimizing a per-pixel loss between the predicted and ground-truth density maps, with the final count obtained by summing the pixel values of the predicted density maps.}
    \label{fig:few-shot-counting-teaser}
\end{figure*}

Formally, 
%in standard density-based counting,
we assume to have a collection of $N$ annotated images denoted as $\mathcal{X} = {(I_1, P_1), \dots, (I_N, P_N)}$, where $I_i$ is the $i$-th image and $P_i \in \mathbb{R}^{2}$ the set of 2D point coordinates roughly localizing the objects to be counted in image $I_i$. The goal is to learn a regression model $f_\theta$ producing a density map $D^{map} = f_\theta(I) \in \mathbb{R}^{H \times W}$. The number of objects $n$ in an image sub-region $S \subseteq I$ is estimated by integrating $D^{map}$ over $S$, i.e., summing up pixel values in the considered region, $n = \sum_{p \in S} D^{map}_p$. \csname rev1\endcsname{The regression model is typically trained by minimizing per-pixel losses, such as the mean squared error between predicted and ground truth density maps.} \csname rev1\endcsname{In the few-shot (or reference-based) counting setting, each image $I_i$ is paired with $K$ exemplars, defined as bounding boxes $B^E = {b_i}_{i=1:K} \in \mathbb{R}^4$ that depict instances of the target class within the same image~\cite{DBLP:conf/cvpr/RanjanSNH21}, with $K=3$ in the original formulation. Object classes are split into base $C_b$ and novel $C_n$ categories, with $C_b \cap C_n = \emptyset$. For $C_b$, both density maps and exemplars are available; for $C_n$, only exemplars are provided, and the goal is to count instances by transferring knowledge from $C_b$ (see also Fig.~\ref{fig:few-shot-counting-teaser}). In contrast, reference-less and text-guided zero-shot settings omit exemplars: the former automatically infers dominant class(es) via instance repetition, while the latter uses a textual description $T_i$ to specify the target class for each image $I_i$.}

\subsection{Reference-based Class-agnostic Counting}
\label{sec:sec:reference-based-works}

\paragraph{FamNet} \csname rev1\endcsname{The seminal work on reference-based object counting is by Ranjan et al.~\cite{DBLP:conf/cvpr/RanjanSNH21}, who first formulated counting as a few-shot regression task. Since the task was novel, the authors also introduced the first dataset tailored for it, a collection of images containing objects belonging to 147 categories named FSC-147; labels comprise dots for the training stage, and exemplar bounding boxes that are used both in the supervised learning and at inference time. See also Sec.~\ref{sec:dataset} for more details.
Finally, the authors proposed the Few-Shot Adaptation and Matching Network (FamNet), a novel architecture designed to address this task. Specifically, it includes two main modules: (i) a multi-scale feature extraction module and (ii) a density prediction module.}
The first consists of the first four blocks picked up from an ImageNet pre-trained ResNet-50 backbone~\cite{DBLP:conf/cvpr/HeZRS16}, frozen during training. The convolutional feature maps produced by the third and fourth blocks provide the representation of the images $I_i$; furthermore, they are also exploited to obtain the multi-scale features of the exemplar images obtained from $B^E$ by performing ROI pooling~\cite{DBLP:conf/iccv/Girshick15} at different scales. Rather, the density prediction module consists of some stacked convolution blocks and upsampling layers; the last one is a $1\times1$ convolutional layer in charge of predicting the final $2D$ density map $D^{map}$ having the same size as the input image. The input of this second module does not correspond to the features produced by the feature extractor, as usual in the class-specific counting task. Instead, the authors computed the convolution operation between the scaled exemplar features and the image features, obtaining multiple correlation maps, one for each scale, that are then concatenated and fed to the density predictor. The reason behind this is making the density regressor agnostic to the visual categories. During the learning stage, FamNet is trained by minimizing the mean squared error (MSE) between the predicted and the ground truth density map. Finally, this work also proposed a sort of fine-tuning that exploits test data to improve performance further. Specifically, it introduced an adaptation loss that exploits the locations of the exemplars included in the test set to be used only at inference time. It comprises two loss components: (i) the Min-Count loss, ensuring that the density sum within each exemplar box is $\geq$ 1, and (ii) the perturbation loss, encouraging Gaussian-shaped density around exemplars---similar to correlation filters used in tracking~\cite{DBLP:journals/pami/HenriquesC0B15,8833463,DBLP:conf/cvpr/Wang0BHT19}.

\paragraph{CFOCNet} Roughly simultaneously, Yang et al. introduced a similar solution~\cite{DBLP:conf/wacv/YangSHC21} named Class-agnostic Few-shot Object Counting Network (CFOCNet). The architecture follows the scheme of FamNet, with an encoder and decoder. The encoder includes two ResNet-50 streams for feature extraction: one for the query image and one for the exemplars (here referred to as reference images). The features concerning the exemplars are matched by computing the convolution operation with the output of a self-attention mechanism~\cite{DBLP:conf/icml/ZhangGMO19} applied to the features relating to the query image. These resulting matching maps are finally fused by exploiting a scale-aware fusing mechanism, and the output is fed into the decoder to compute the density map. The decoder is similar to the one exploited in FamNet---some stacked convolution blocks and upsampling layers---and it is supervised by standard MSE loss as well as by SSIM loss~\cite{DBLP:journals/tip/WangBSS04} to catch local pattern consistency~\cite{DBLP:journals/tnn/ChenZZZK24}.

\paragraph{VCN} \csname rev1\endcsname{Some of the authors of~\cite{DBLP:conf/cvpr/RanjanSNH21} extended FamNet in~\cite{DBLP:conf/cvpr/RanjanH22}, addressing its limitations due to the small dataset used for supervised training. They introduced the Vicinal Counting Network (VCN), which retains the core of FamNet with minor changes and incorporates a generator to augment and synthesize images from the vicinity of training samples, expanding the dataset.} Specifically, the generator is fed with an input image $I \in \mathbb{R}^{H \times W \times 3}$ and a noise vector $z \in \mathbb{R}^m$, and it produces an augmented version $I_z \in \mathbb{R}^{H \times W \times 3}$ of $I$. This augmented image serves as the input to the feature extractor and the subsequent regressor module, together with the original input $I$. The entire system is then trained jointly in a cooperative fashion. \csname rev1\endcsname{However, unlike the min-max game typical of GANs~\cite{NIPS2014_5ca3e9b1}, VCN relies on a standard minimization problem, combining MSE loss for density regression with a reconstruction loss to enforce similarity between generated and input samples, and a diversity loss to prevent identity mapping from $I$ to $I_z$.} Finally, the authors used the same test time adaptation loss of FamNet~\cite{DBLP:conf/cvpr/RanjanSNH21}.

\paragraph{RCAC} \csname rev1\endcsname{Another work inspired by FamNet is~\cite{DBLP:conf/eccv/GongZ0DS22}. First, the authors explored the main limitations of FamNet by analyzing failure cases, identifying a strong dependence on exemplar diversity---exemplar boxes are few and subjectively annotated, and they may not cover all object instances or ensure intra-class diversity. To address this, they proposed the Robust Class-Agnostic Counter (RCAC), which uses a two-stream architecture: one branch processes RGB images $I_i$ with a FamNet-like feature extractor, while the other uses a VGG-based extractor on edge images derived from $I_i$ throughout an edge detector.} This second feature extractor is randomly initialized and, different from the RGB feature extractor, optimized during training. The reason claimed by the authors behind using edge images is that edges are a sort of class-agnostic knowledge, i.e., shape is a more reliable cue across instances than color. \csname rev1\endcsname{Then, ROI pooling is applied to both branches to crop exemplar features, which are then enhanced via a feature augmentation module. After that, feature correlation is obtained by computing the convolution operation between the latter and the full feature maps, like in previous works, and the density map is predicted using the same module as in FamNet.} In detail, the feature augmentation module aims to generate more exemplars of different colors, shapes, and scales in the feature space by combining the provided exemplars, sharing the goal of VCN~\cite{DBLP:conf/cvpr/RanjanH22} to expand the training set.

\paragraph{BMNet+} Another notable work that represents a milestone is~\cite{DBLP:conf/cvpr/Shi0FL022}. Here, Shi et al. argued that two factors played crucial roles in previous works: (i) feature representation of query images and exemplars, and (ii) similarity metric for matching exemplar and image features. Previous methods either exploited a fixed~\cite{DBLP:conf/cvpr/RanjanSNH21,DBLP:conf/cvpr/RanjanH22} or a learnable~\cite{DBLP:conf/wacv/YangSHC21} feature extractor but employed a similarity metric relying on the fixed inner product~\cite{DBLP:conf/cvpr/RanjanSNH21,DBLP:conf/cvpr/RanjanH22,DBLP:conf/wacv/YangSHC21,DBLP:conf/eccv/GongZ0DS22, DBLP:journals/prl/McCarthyVOTAH23}, which yielded insufficient matching outcomes. Hence, the authors proposed the Bilinear Matching Network (BMNet+) to improve the similarity modeling. Specifically, the authors designed a learnable bilinear similarity loss to supervise the similarity-matching results, thus going beyond the fixed inner product. The authors took inspiration from metric learning, which aims to embed data into a space where similar samples are pulled close while dissimilar ones are pushed away~\cite{DBLP:conf/eccv/MusgraveBL20}. In more detail, their similarity loss pulls close the features between the exemplar and target instances while pushing away the features between exemplars and background patches. %Furthermore, similar to~\cite{DBLP:conf/wacv/YangSHC21}, the authors also proposed a self-attention mechanism to represent self-similarity among features. 

\paragraph{SPDCN} Lin et al.~\cite{DBLP:conf/bmvc/LinYM0LLHYC22} focused on the robustness of exemplar feature representations, such as in VCN~\cite{DBLP:conf/cvpr/RanjanH22} and RCAC~\cite{DBLP:conf/eccv/GongZ0DS22}, and proposed the Scale-Prior Deformable Convolution Network (SPDCN). \csname rev1\endcsname{The authors argued that most prior works often overlooked or poorly handled the exemplar scale during feature extraction. To address this, they leveraged size information from exemplars by exploiting scale-prior deformable convolutions~\cite{DBLP:conf/iccv/DaiQXLZHW17,DBLP:conf/cvpr/ZhuHLD19} to discard features inconsistent with scale, enabling more meaningful feature extraction.} %Specifically, they considered a scale-prior backbone consisting of the first ten convolutional layers of an ImageNet pre-trained VGG-19 network~\cite{Simonyan15}, where some specific layers have been converted to scale-prior deformable convolutions~\cite{DBLP:conf/iccv/DaiQXLZHW17,DBLP:conf/cvpr/ZhuHLD19} to exploit the critical scale information from the exemplars. 
%More in detail, exemplars' scale embeddings are computed by exploiting the exemplars' scales, and the receptive fields of the deformable convolutions are adjusted accordingly. 
Afterward, these semantic features are split into two branches for segmentation and density estimation, respectively. In the former branch, such as in FamNet and other previous works, ROI pooling~\cite{DBLP:conf/iccv/Girshick15} is exploited to extract feature maps concerning the exemplars; next, these features are averaged into a representation vector, and a whole similarity map is calculated by computing the cosine similarity between this vector and the image feature map. In the density estimation branch, class-agnostic features are calculated using element-wise multiplication like in~\cite{DBLP:conf/cvpr/RanjanSNH21,DBLP:conf/cvpr/RanjanH22,DBLP:conf/wacv/YangSHC21,DBLP:conf/eccv/GongZ0DS22} between the similarity map and the image feature map. Then, they are fed into a decoder that follows the design of the Pixel Shuffle Crowd Counter (PSCC)~\cite{DBLP:journals/tcyb/WangLGL22}, which employs pixel-shuffling~\cite{DBLP:conf/cvpr/ShiCHTABRW16} for the upsampling operations, and that is responsible for predicting the final density map. %Finally, another novel contribution of this paper is the usage of the generalized loss~\cite{DBLP:conf/cvpr/WanLC21} to optimize the density predictor. Unlike the MSE loss, utilized in most previous works, generalized loss measures the distance between the predicted density map directly against the dot map, and not the ground truth density map generated by convolving Gaussian kernels. In this way, the cost function is adjusted accordingly to the given exemplars, and it is not fixed to the object-specific sizes.

\paragraph{MACnet} The authors in~\cite{DBLP:journals/prl/McCarthyVOTAH23} argued that bounding boxes are not the ideal representations for most exemplars as they often capture undesirable background portions that harm the matching process. Therefore, they proposed exploiting segmentation masks as an alternative, introducing the Mask Augmented Counting network (MACnet). In more detail, the proposed architecture consists of three modules. First, an ImageNet~\cite{DBLP:journals/ijcv/RussakovskyDSKS15} pre-trained ResNet-50~\cite{DBLP:conf/cvpr/HeZRS16} is used as a feature extractor to embed both the query and the exemplars into a feature space. Specifically, its input consists of the query image $I$, the exemplars' bounding box coordinates $B^E$, and their associated masked counterparts $B_{m}^{E}$ computed with the pre-trained Deep Extreme Cut network (DEXTR)~\cite{DBLP:conf/cvpr/ManinisCPG18}. %However, unlike previous works, the exemplars are not retrieved with bounding boxes but through their extreme points. The reason claimed by the authors is twofold: (i) annotation with extreme points is approximately four times faster than using bounding boxes~\cite{DBLP:conf/cvpr/ZhouZK19}, and (ii) extreme points are exploited to compute the segmentation masks with a pre-trained Deep Extreme Cut network (DEXTR)~\cite{DBLP:conf/cvpr/ManinisCPG18}. It is also worth noting that while previous literature got the exemplars by cropping them in the feature space, here they are retrieved in the pixel space since elements in the extracted feature tensor already capture background information due to the large receptive field of the feature extractors. 
The second module consists of a residual layer fed with the feature representation of the exemplars and their masked counterparts. Specifically, it mitigates errors during segmentation mask creation by learning from residual information instead of immediately discarding background information. Finally, the third module is responsible for predicting the density map, supervised by the MSE loss. As in previous papers~\cite{DBLP:conf/cvpr/RanjanSNH21,DBLP:conf/wacv/YangSHC21}, it is fed with a similarity map obtained by convolving the feature map of the query image with the feature map resulting from the residual module as its filter, i.e., by performing the inner product between these two feature maps. However, here, the final count is derived not directly from integrating the density map but by counting the peaks utilizing a peak-finding algorithm. A final observation concerns the usage of a test time adaptation loss like in~\cite{DBLP:conf/cvpr/RanjanSNH21,DBLP:conf/cvpr/RanjanH22}, aiming to encourage the network in distinguishing background from foreground in the provided exemplars.

\paragraph{SAFECount} You et al. introduced SAFECount in~\cite{DBLP:conf/wacv/YouYLLCL23}, which uses Similarity-Aware Feature Enhancement blocks. These blocks are composed of two main modules: (i) a Similarity Comparison Module (SCM), and (ii) a Feature Enhancement Module (FEM). In SCM, query and reference features extracted by an ImageNet pre-trained ResNet-18~\cite{DBLP:conf/cvpr/HeZRS16} are compared in a learnable projection space, and a score map is obtained for each reference image. The similarity map is obtained from the score map by normalizing the exemplar dimension and spatial dimension. The FEM uses the similarity map to weight the reference features, highlighting regions in the query image that are similar to the reference images. These weighted features are then fused into the query features, creating an enhanced representation that emphasizes the areas containing objects of interest. 

\paragraph{LOCA} Djukic et al. focused on the strength of exemplar feature representations, arguing that, ideally, the prototypes should generalize over the appearance of the selected object category in the image. They introduced the Low-shot Object Counting network with iterative prototype Adaptation (LOCA)~\cite{DBLP:conf/iccv/EukicLZK23}. \csname rev1\endcsname{The core of the method is an object prototype extraction module that separately encodes shape and appearance from exemplars and refines them iteratively. Appearance is injected by initializing prototypes with RoI-pooled features from exemplar boxes over a ResNet50-based query feature map. However, this pooling makes the queries shape-agnostic, so shape information is also injected using exemplar width and height. Then, the final prototypes are obtained via an iterative adaptation module with cross-attention blocks.}
%The core of their solution is the object prototype extraction module, which separately analyzes the shape and appearance characteristics of exemplars and iteratively refines them into object prototypes. Specifically, appearance information is injected by initializing the prototypes with image features computed by RoI pooling~\cite{DBLP:conf/iccv/Girshick15} exemplar bounding boxes over a query image feature map obtained with a pre-trained ResNet50~\cite{DBLP:conf/cvpr/HeZRS16}. However, the pooling operation makes the appearance queries shape-agnostic, since it maps features from different spatial shapes into rectangular queries of the same size. Thus, the authors also injected shape information by initializing the prototypes with exemplar width and height features. Finally, the shape and appearance queries are converted into object prototypes by an iterative adaptation module using a recursive sequence of cross-attention blocks.

\paragraph{ConCoNet} 
Soliven et al. introduced the Contrast Counting Network (ConCoNet) in~\cite{DBLP:journals/prl/SolivenVOTAH23}. They proposed to use not only the (positive) exemplars as in previous works but also the negative exemplars, i.e., object instances defining what \textit{not} to count. Their contribution can be incorporated into other class-agnostic counting models, and, by way of examples, they reported the results obtained using two of the milestones for the reference-based class-agnostic counting task, e.g., FamNet~\cite{DBLP:conf/cvpr/RanjanSNH21} and BMNet+~\cite{DBLP:conf/cvpr/Shi0FL022}. Specifically, the class-agnostic similarity maps are computed not only for the query image and positive exemplar pairs but also for the negative ones; the final similarity map input to the density regressor is obtained by simply subtracting the negative similarity map from the positive one. Since CAC datasets do not contain negative exemplars, the authors derived them under the assumption that users provide negative examples during inference, thus performing a sort of adaptation at test time; instead, to enable the training, they selected bounding boxes randomly from the query image regions classified as background by both the ground-truth and the predicted density maps.

\paragraph{SSD} Instead, in~\cite{DBLP:conf/ijcai/Xu0Z24}, the authors leveraged distinct similarity distribution characteristics between several parts of exemplar objects. They argued that different parts of an exemplar object generate distinct similarity distribution patterns. For instance, the similarity distribution at the center of the object gradually decreases from the center toward the outer regions, while the distribution at the edges varies depending on the location. To capture this, they proposed the Spatial Similarity Distribution (SSD) network, which produces a 4D similarity tensor. This allows for flexible extraction of point-to-point similarity distribution information between query and exemplar features through convolution operations in 4D space.

\paragraph{SATCount} Differently, Wang et al. proposed a Scale-Aware Transformer-based class-agnostic counting framework named SATCount~\cite{DBLP:journals/nn/WangYWLC24}. Here, to combine the features from exemplars and the query, the authors developed a feature fusion module comprising two sub-modules: the Scale-Aware Module (SAM) and the Similarity Modeling Module (SMM). SAM combines visual and scale information to generate scale-aware visual features of exemplar objects. Meanwhile, SMM uses cross-attention to model the similarity between exemplar and query features, allowing for long-distance feature associations and the integration of global information from both exemplars and queries.

\paragraph{CSTrans} Gao and Huang~\cite{DBLP:journals/pr/GaoH24} drew inspiration from FamNet, enhancing its architecture with two new modules: Correlation-guided Self-Activation (CSA) and Local Dependency Transformer (LDT). The resulting approach, named Correlation-guided Self-Activation Transformer (CSTrans), improves the correlation representation between the exemplar patch and query image features. Additionally, it incorporates a density map predictor based on Transformers~\cite{DBLP:conf/nips/VaswaniSPUJGKP17} that is fully learnable and better equipped to model local context dependencies.

\paragraph{CountVers} The authors in~\cite{DBLP:journals/kbs/YangCDWZ24} focused on a novel correlation mechanism between exemplar-query features, overcoming the limitations of kernel-wise convolution characterizing previous works such as FamNet. \csname rev1\endcsname{Specifically, they proposed CountVers, which integrates both channel-wise and spatial-wise correlation to handle small and large object counting. The method builds on two insights: small objects have localized features, making channel-wise correlation beneficial, while large objects span broader regions with complex spatial semantics, requiring spatial-wise correlation for accurate counting.}

\paragraph{CACViT} 
\csname rev1\endcsname{Recently, Zhicheng et al.~\cite{DBLP:conf/aaai/WangX0024} introduced a framework based on a pre-trained Vision Transformer (ViT) and an "extract-and-match" paradigm that removes the need for multiple feature extractors or post-processing. The attention mechanism in ViT is shown to handle both feature extraction and matching by processing concatenated query and exemplar tokens. Within the ViT, self-attention extracts features from both inputs, while cross-attention enables matching between them.}

%Zhicheng et al.~\cite{DBLP:conf/aaai/WangX0024} introduces a unified extract-and-match framework for counting objects in images regardless of class, using Vision Transformers (ViTs). Unlike traditional methods that separate feature extraction and matching, CACViT processes query and exemplar images simultaneously within a single ViT architecture. The query image and exemplars are divided into patches, serving as tokens for the ViT, which processes these through multiple layers comprising LayerNorm, Multi-Head Self-Attention (MHA), and Multi-Layer Perceptron (MLP). The self-attention mechanism in the ViT captures relationships between different tokens, with the attention map \(A\) decoupled into four sub-maps: \(A_{\text{query}}\) (self-attention within the query), \(A_{\text{class}}\) (query to exemplar attention), \(A_{\text{match}}\) (cross-attention between query and exemplar), and \(A_{\text{exp}}\) (self-attention within exemplars). This approach is enhanced by Scale Embedding (SE) and Magnitude Embedding (ME) to compensate for information loss due to resizing and normalization. The output is generated through a regression decoder that produces a density map, visually representing object counts. The self-attention is formulated as \(\text{Attention}(Q, K, V) = \text{softmax}\left(\frac{QK^T}{\sqrt{D}}\right)V\), where \(Q, K, V\) are derived from the input tokens. This novel framework significantly simplifies the counting process and improves accuracy, setting a new state-of-the-art for class-agnostic counting.

\paragraph{DAVE} 
Pelhan et al. recently introduced a detect-and-verify paradigm (DAVE)~\cite{DBLP:conf/cvpr/PelhanLZK24} that combines the strengths of density-based methods---best suited for crowded scenarios---with those of detection-based techniques, which provide precise object localization. This approach follows a two-stage pipeline. In the first detection stage, DAVE generates a high-recall set of candidate bounding boxes by first estimating the object centers through a location density map and then predicting bounding box parameters using a regression head.
The verification stage refines these detections by removing false positives. Each candidate is analyzed based on its appearance features, which are compared to features of the provided exemplars using clustering methods. Detections that do not align with exemplar clusters are classified as outliers and removed. The refined bounding boxes are then used to update the density map, which further improves count estimation accuracy.

\paragraph{TFPOC} Different from the above-described architectures, TFPOC~\cite{DBLP:conf/wacv/Shi0Z24} employed a detection-based technique, leveraging the popular Segment Anything Model~\cite{DBLP:conf/iccv/KirillovMRMRGXW23} for instance segmentation that does not require additional training or fine-tuning for class-agnostic counting. Initially, the authors presented a baseline method. This involves generating binary masks for reference objects using input prompts such as points or boxes, computing a similarity map between the image features and reference object features via cosine similarity, and producing masks for all objects in the image using a grid of point prompts. Target objects are identified by calculating similarity scores for each mask and applying a threshold to determine relevance. The final count is obtained by summing the identified target objects. However, this vanilla approach is computationally intensive as it processes all objects in the image and struggles with setting an optimal similarity threshold, which can lead to undercounting or overcounting. To address these limitations, the authors proposed an advanced prior-guided mask generation method, enhancing the segmentation process with three types of priors. The similarity prior uses a binarized similarity map to guide the segmentation model in focusing on target regions. The segment prior prevents redundant processing by maintaining an overall segment map and refining segmentation iteratively. The semantic prior incorporates reference object features to better guide the mask decoder in identifying and segmenting the target objects.

\paragraph{UPC} A token-based framework for unified open-world text-guided counting (UPC) is introduced in~\cite{DBLP:conf/aaai/0018C24}. Prompts, which may take the form of boxes, points, or text, are first transformed into a unified representation called a prompt mask. For box and point prompts, this mask is derived directly from the labeled regions or pixels within the image. As in previous methods, image features are extracted using a CNN-based encoder, and a cross-attention mechanism is applied to determine the similarity between the prompt token and the image features. This similarity guides the creation of density features, which are subsequently decoded by a CNN to produce the final density map. \csname rev1\endcsname{However, in this work, to improve the robustness of the model, the predicted density map is iteratively refined by reusing it as a new prompt mask, leading to convergence toward consistent outputs. A fixed-point loss function, based on implicit differentiation, stabilizes training and improves parameter optimization. Additionally, a contrastive training strategy is used: positive samples contain objects matching the prompt, while negative ones do not. This forces the model to produce accurate density maps for positives and zero-density maps for negatives, reducing errors from noisy or ambiguous prompts.}

\paragraph{PseCo} As in TFPOC~\cite{DBLP:conf/wacv/Shi0Z24}, the authors in~\cite{DBLP:conf/cvpr/HuangD0ZS24} exploited a detection-based approach relying on the popular SAM~\cite{DBLP:conf/iccv/KirillovMRMRGXW23} for instance segmentation. Specifically, they introduced a framework termed Point, Segment and Count (PseCo). Different from TFPOC, PseCo does not use uniform grid points to prompt SAM to segment all objects.
Indeed, the authors argued that using grid points may fall short in capturing all objects, particularly in densely crowded scenes, leading to the omission of many small objects. \csname rev1\endcsname{Although increasing the density of grid points could help address this limitation, it would result in substantial computational overhead. Thus, a class-agnostic object localization method was proposed in~\cite{DBLP:conf/cvpr/HuangD0ZS24}, estimating a heatmap from which object coordinates are inferred and used to guide SAM. then introduced a second stage that integrates another foundation model, the Contrastive Language–Image Pretraining (CLIP) model~\cite{DBLP:conf/icml/RadfordKHRGASAM21}, to classify regions identified in the first stage based on given exemplars.}

\paragraph{\textcolor{black}{CountDiff}}\csname rev4round3\endcsname{Hui et al.~\cite{DBLP:conf/eccv/HuiWRL24} proposed CountDiff, a framework designed to boost CAC by leveraging knowledge from a pre-trained diffusion model. To encode exemplar-specific information, CountDiff learns an object-specific embedding by mapping image features into the text embedding space, enabling the diffusion model to interpret the visual characteristics of the target object.
To enhance generalization in CAC, CountDiff complements this with an object-agnostic embedding, capturing transferable knowledge from the diffusion model’s cross-attention layers. By combining both embeddings, the framework effectively exploits cross-attention maps to identify regions relevant for CAC.
Finally, a lightweight test-time adaptation step fine-tunes the object-specific embedding using the provided exemplars, further boosting accuracy on novel objects.}

%\paragraph{\textcolor{darkgreen}{GeCo}} \csname rev4round3\endcsname{~\cite{DBLP:conf/nips/PelhanLZK24} TODO.}

%\subsubsection*{CFENet}~\cite{DBLP:journals/ivc/ZhangZCWH25}

%\subsubsection*{MEAMNet}~\cite{DBLP:journals/prl/ZhangHZLCL24}

\subsection{Reference-less Class-agnostic Counting}
\label{sec:sec:reference-less-works}

\paragraph{RepRPN-Counter} The first reference-less class-agnostic counting method was introduced in~\cite{DBLP:conf/accv/RanjanN22}. In this paper, Ranjan et al. introduced the Repetitive Region Proposal Network Counter (RepRPN-Counter), which predicts separate density maps for each repetitive object category in a given query image. The model is based on FamNet~\cite{DBLP:conf/cvpr/RanjanSNH21} and includes an upstream module called the Repetitive Region Proposal Network (RepRPN), designed to automatically identify exemplars of repetitive objects in the image, which are then passed to FamNet for further processing. RepRPN takes inspiration from the Region Proposal Network (RPN) of the popular Faster R-CNN object detector~\cite{DBLP:conf/nips/RenHGS15}. RPN typically calculates proposal boxes and objectness scores, but RepRPN additionally outputs a so-called repetition score, indicating how frequently an object within the proposals appears in the image.
%The authors provided an illustrative example in their paper. Suppose to have a query image containing \textit{m} cats and \textit{n} oranges: the RepRPN should predict, for any cat proposal, the repetition score \textit{m}, and, in the same way, the repetition score \textit{n} for any orange proposal. 
These repetition scores are used to select exemplars from the repetitive object categories in the query image, with the highest-scoring proposals chosen as exemplars. 
%However, the original RPN generated region proposals based on anchors---predefined bounding boxes of varying sizes and aspect ratios used as reference points for object detection---which did not cover the entire image, a critical requirement for computing the global repetition score. To address this, the authors incorporated Encoder Self-Attention layers~\cite{DBLP:conf/nips/VaswaniSPUJGKP17}, which can integrate information from similar feature vectors across the entire convolutional feature map of the image, enabling efficient computation of the global repetition score at any anchor location. 
\csname rev1\endcsname{A key challenge in training RepRPN was the lack of labels in class-agnostic counting datasets like FSC-147~\cite{DBLP:conf/cvpr/RanjanSNH21}, where annotations are provided for only one object category, despite the presence of other unannotated objects (see also Sec.~\ref{sec:dataset}). Penalizing predictions for these unlabeled objects could harm performance. To overcome this, the authors used knowledge transfer: RepRPN trained on MSCOCO~\cite{DBLP:conf/eccv/LinMBHPRDZ14} acted as a teacher to assign objectness and repetition scores to proposals not overlapping with annotated objects, while a pre-trained FamNet generated density maps for these proposals using single exemplars.}

\paragraph{RCC}~\cite{DBLP:journals/corr/abs-2205-10203} went a step further by proposing RCC - Reference-less class-agnostic counting. Unlike previous works, it did not use point-level supervision---besides not using any exemplars as in the reference-less setting---thus proposing a weakly-supervised methodology. Specifically, their proposed network directly regresses to a single scalar count prediction for the entire input image, thus not requiring ground-truth density maps for supervised learning. To create a general and informative feature space without labels, the authors implemented a self-supervised knowledge distillation technique based on~\cite{DBLP:conf/iccv/CaronTMJMBJ21}. They encouraged the learning of meaningful image representations by fostering agreement between a fixed teacher network, which processes large global image crops, and a student network, which analyzes smaller local image crops. This alignment is achieved by minimizing the cross-entropy between the probability distributions produced by both networks.

\paragraph{CounTR} Another milestone concerning reference-less class-agnostic counting is represented by Counting TRansformer - CounTR~\cite{DBLP:conf/bmvc/LiuZZX22} (pronounced "counter"). In this work, Liu et al. built on the idea that self-similarity serves as a strong prior in visual object counting. They introduced a transformer-based architecture where self-similarity can be explicitly captured by the inherent attention mechanisms, both between input image patches and with the exemplars (if any). Specifically, this architecture features two visual encoders: (i) a Video Transformer (ViT)~\cite{DBLP:conf/iclr/DosovitskiyB0WZ21}, which maps input image features into a high-dimensional feature space, and (ii) a lightweight CNN, responsible for extracting visual features from the exemplars (if available). The subsequent feature interaction module is responsible for merging information from both encoders. It consists of a series of standard transformer decoder layers, where the image features serve as the \textit{Query}, and two separate linear projections of the exemplar features, or a learnable special token if no exemplars are present, act as the \textit{Key} and \textit{Value}. This architecture aligns well with the self-similarity prior in counting tasks: self-attention captures similarities between image regions, while the cross-attention between \textit{Query} and \textit{Value} compares these regions with the exemplars or learns to bypass the CNN branch when using the learnable token, i.e., in the reference-less setting where no exemplars are provided. Additionally, the authors proposed a two-stage training approach, where the transformer-based image encoder is first pre-trained using self-supervision through masked image modeling~\cite{DBLP:conf/cvpr/HeCXLDG22}, followed by supervised fine-tuning for the counting task. %Finally, another notable contribution tackled the long-tailed issue in existing counting datasets, where only a few images contain a large number of instances. To handle this, they introduced a scalable pipeline for synthesizing training images with large object counts by cropping random-sized square regions from existing images and arranging them together in a collage. 

\paragraph{GCNet} Also the authors of~\cite{DBLP:journals/pr/WangLZTG24} introduced a technique that does not use point-level supervision such as~\cite{DBLP:journals/corr/abs-2205-10203}, named GCNet - Generalized Counting Network. However, differently from~\cite{DBLP:journals/corr/abs-2205-10203}, GCNet uses a pseudo-Siamese structure to capture pseudo-exemplar cues from the resized raw image. Specifically, the main branch of the network extracts primary 2D feature maps from an input image $I_i$. In contrast, the secondary branch is in charge of generating exemplar-focused outputs by exploiting a methodology inspired by vision transformers~\cite{DBLP:conf/iclr/DosovitskiyB0WZ21}.
%: (i) a sequence of overlapping patches is extracted using the unfold operation over the 2D feature maps computed by the main branch, (ii) pseudo-exemplar patches are computed using a per-pixel average operation along the patch dimension, ensuring to capture the common attributes across local regions, (iii) pseudo-exemplar patches are divided into 16 sub-patches of size $2 \times 2$, and (iv) each sub-patch is flattened into a 1D token and then transformed through linear projection. 
The output of the main branch is then further processed through an anisotropic encoder to capture discriminative features along three directions in feature space---horizontal, vertical, and basis/channel directions. Finally, these two outputs are fed into a dual-attention condenser. The same authors also proposed a reference-based version of GCNet, which attaches an additional ResNet-50 backbone to the front-end of the original GCNet for processing labeled exemplar patches; the exemplar features are then fed into the same back-end structure of GCNet to predict an auxiliary single count.

\paragraph{Adaptations} Some existing reference-based techniques have been adapted to the reference-less setting. Specifically, some minor modifications have been implemented to the module of LOCA~\cite{DBLP:conf/iccv/EukicLZK23} in charge of extracting object prototypes. Moreover, the authors of DAVE~\cite{DBLP:conf/cvpr/PelhanLZK24} made adjustments to their architecture, replacing the location density prediction component with a reference-less version of LOCA. However, the detection stage and most of the verification stage remained unchanged. The only alteration in the verification stage involves the cluster selection process: clusters with a size of at least 45\% of the largest cluster are retained as positive detections, while the others are classified as outliers. This adjustment addresses potential cluster fragmentation due to the lack of exemplars that define appearance similarity levels. Lastly, the authors of~\cite{DBLP:conf/accv/RanjanN22} proposed a modified version of FamNet, incorporating a standard Region Proposal Network (RPN) from the widely used Faster R-CNN object detector~\cite{DBLP:conf/nips/RenHGS15} for exemplar generation. 
%\csname rev4round3\endcsname{GeCo~\cite{DBLP:conf/nips/PelhanLZK24} TODO.}

%\subsection{Open-world Text-specified Object Counting}
\subsection{Open-world Text-guided Object Counting}
\label{sec:sec:prompt-based-works}

%Most of the state-of-the-art prompt-based CAC approaches rely on vision-language foundation models that map text-image pairs to a joint embedding space; the produced features are then passed to a decoder in charge of producing density maps. 

\paragraph{ZSC} \csname rev1round2\endcsname{Xu et al.~\cite{DBLP:conf/cvpr/XuL0RS23} were the first to propose a open-world text-guided approach, introducing a more versatile paradigm called zero-shot counting (ZSC) where users can specify textual descriptions of the object classes they want to count, representing a sort of trade-off between previous class-agnostic counting paradigms---humans are still in the loop but can specify classes with less effort and in a more natural way.}
\csname rev1\endcsname{In more detail, their proposal is grounded on an effective strategy for selecting meaningful patches from the input image that contain the target object specified by a class name, to be used as exemplars. To this end, they construct a class prototype in a pre-trained embedding space using the provided class name, and then select patches whose embeddings are among the $k$-nearest neighbors to that prototype. This is implemented using a conditional variational autoencoder (VAE) built on the widely adopted Contrastive Language-Image Pretraining (CLIP) model~\cite{DBLP:conf/icml/RadfordKHRGASAM21}, which generates visual exemplar prototypes conditioned on the semantic embedding of the given category name.}

\paragraph{CounTX} Concurrently, Amini-Naieni et al.~\cite{AminiNaieni23} proposed an alternative solution called CounTX (pronounced "Count-text"). CounTX can be trained end-to-end and accepts more detailed specifications of target objects for counting, rather than relying solely on class names. To support this, the authors introduced an enhanced version of the FSC-147 dataset~\cite{DBLP:conf/cvpr/RanjanSNH21}, named FSC-147-D, where object classes are described using more fine-grained and structured natural language sentences (See also Sec.~\ref{sec:dataset} for more details).  Specifically, CounTX extends the reference-less CounTR architecture~\cite{DBLP:conf/bmvc/LiuZZX22} to the open-world text-guided setting. Like CounTR, it features a transformer-based image encoder, but it utilizes the CLIP vision transformer B-16~\cite{DBLP:conf/icml/RadfordKHRGASAM21}, pre-trained in a contrastive manner with a transformer-based text encoder using image-text pairs from LAION-2B~\cite{DBLP:conf/nips/SchuhmannBVGWCC22}. In more detail, the image encoder encodes each input image into a spatial map of 512-dimensional feature vectors, while the text encoder processes the class description into a single 512-dimensional feature vector. To combine these features, the authors employed a transformer-based feature interaction module, similar to CounTR, but instead of matching image patches to visual exemplars, in CounTX, it is exploited to compute the similarities between image patches and class descriptions. Finally, the output of the feature interaction module is reshaped to a spatial feature map and fed into a progressive four-block convolution and upsampling operation to produce density maps.

\paragraph{CLIP-Count} CLIP-Count~\cite{DBLP:conf/mm/JiangLC23} is another method built on the Contrastive Language–Image Pretraining (CLIP) model~\cite{DBLP:conf/icml/RadfordKHRGASAM21}, designed to align text embeddings with dense visual features. In this approach, text embeddings are compared with image patch embeddings. Key features of this method include: (i) the use of an InfoNCE-based contrastive loss~\cite{DBLP:journals/corr/abs-1807-03748}, commonly employed in contrastive learning to encourage meaningful and discriminative representations by pulling similar data points (positive pairs) closer in the latent space and pushing dissimilar ones (negative pairs) further apart. This loss is used alongside the classic MSE loss, applied to binary ground-truth masks that distinguish between positive and negative patch regions; and (ii) a hierarchical patch-text interaction module that propagates semantic information across different resolution levels of visual features. For density map generation, the decoder employs a conventional design based on convolutional and upsampling layers.

\paragraph{VLCounter} In~\cite{DBLP:conf/aaai/KangMKH24}, the authors incorporated three modules to efficiently fine-tune CLIP for the counting task and to exploit intermediate features across different encoding layers of CLIP in the decoding stage. In particular: (i) a semantic-conditioned prompt-tuning module is responsible for fine-tuning CLIP efficiently, using conditioning via semantic embeddings to generate patch embeddings that highlight the region of interest, rather than relying on simple learnable prompts; (ii) a learnable affine transformation module adjusts the semantic-patch similarity map produced by the previous module to better suit the counting task; (iii) a segment-aware skip connections module is utilized to capture intermediate features from various encoding layers of CLIP, enhancing the generalization ability of the decoder and providing rich contextual information.

\paragraph{GroundingREC}~\cite{DBLP:conf/cvpr/DaiLC24} \csname rev1\endcsname{addressed a key limitation in object counting, which is often restricted to class-level granularity and overlooks fine-grained distinctions. However, in real-world scenarios, counting typically requires contextual or referential input---for example, distinguishing traffic flow directions or separating stationary from moving pedestrians and vehicles at a junction.}
%addressed a key limitation in object counting, where the task is typically confined to the class level and fails to account for fine-grained details within the class. In practical applications, counting often requires contextual or referring human input to target specific objects, such as monitoring traffic flow in different directions or distinguishing between pedestrians and vehicles that are stationary or moving at various points of a junction. 
To tackle this, the authors introduced a novel task called Referring Expression Counting (REC), which focuses on counting objects with distinct attributes within the same class. To evaluate the REC task, they developed a new dataset named REC-8K, comprising 8011 images and 17,122 referring expressions (See also Sec.~\ref{sec:dataset} for more details). The methodology leverages the open-set detector GroundingDino~\cite{DBLP:conf/eccv/LiuZRLZYJLYSZZ24}, adapted with specific modifications such as replacing bounding boxes with box centers and utilizing the CLS token as a representation of the global semantics of the referring expression (rather than individual text tokens). %Additionally, the authors introduced two modules to enhance REC performance: (i) a global-local feature fusion that integrates global image features to infer relational attributes, improving the model's understanding of the context; (ii) a contrastive learning module that is designed to learn discriminative features associated with attributes by processing multiple referring expressions for different attributes of same-class objects. 
%Besides the REC task, the authors also evaluated their method of class-agnostic counting to show that their approach is generalizable to prior tasks.
In addition to REC, they also evaluated their method on class-agnostic counting, demonstrating its generalizability to prior tasks.

\paragraph{VA-Count} \csname rev1\endcsname{In~\cite{DBLP:conf/eccv/ZhuYYGWZH24}, the authors addressed the challenge of identifying high-quality exemplars for zero-shot counting, introducing the Visual Association-based Zero-shot Object Counting (VA-Count) framework. VA-Count includes two main components: the Exemplar Enhancement Module (EEM) and the Noise Suppression Module (NSM), which respectively refine exemplar selection and mitigate the impact of incorrect samples. The EEM leverages GroundingDino~\cite{DBLP:conf/eccv/LiuZRLZYJLYSZZ24}, as in~\cite{DBLP:conf/cvpr/DaiLC24}, but adds a binary filter to select candidate exemplars containing exactly one object. Despite this, errors may occur when selected exemplars do not belong to the target category. The NSM addresses this by identifying and filtering out such negative exemplars. Using contrastive learning, it distinguishes between relevant and irrelevant samples, enhancing the robustness of the counting process.}

\begin{figure*}[!t]
    \centering
    \includegraphics[width=.99\linewidth]{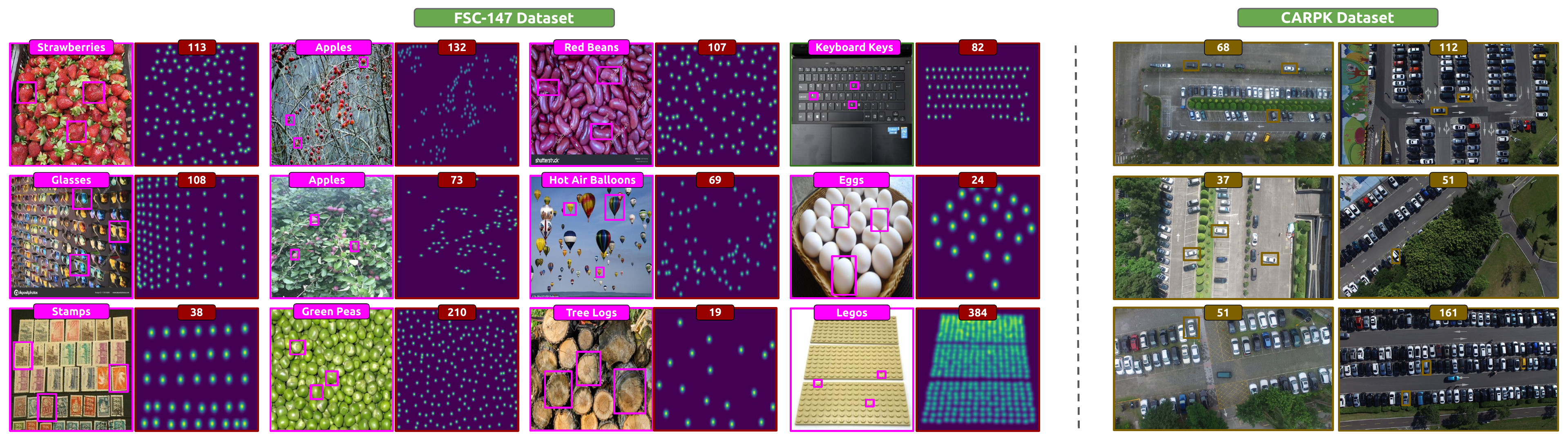}
    \caption{\textbf{Dataset samples.} \csname rev1\endcsname{We show some samples from the CAC gold standard benchmarks, together with bounding boxes marking the exemplars. (a) The FSC-147 dataset~\cite{DBLP:conf/cvpr/RanjanSNH21} includes more than 6,000 images belonging to 147 object classes. We present sample images along with the provided labels, which include density maps and three bounding boxes localizing the exemplars. (b) The CARPK dataset~\cite{DBLP:conf/iccv/HsiehLH17}, a collection of drone-view images tailored for vehicle counting, which is often used to assess cross-dataset generalization capabilities on the specific \textit{vehicle} category. We show sample images together with 12 bounding boxes marking the exemplars (as set by~\cite{DBLP:conf/cvpr/RanjanSNH21}).}}
    \label{fig:dataset}
\end{figure*}

\paragraph{Adaptations} 
Some reference-based techniques have been adapted for the open-world text-guided setting. For example, the authors of DAVE~\cite{DBLP:conf/cvpr/PelhanLZK24} introduced minor adjustments to their architecture, particularly in the cluster selection process during the verification stage. In this approach, the text prompt embedding is derived using the CLIP model~\cite{DBLP:conf/icml/RadfordKHRGASAM21} and compared to the CLIP embeddings of the identified clusters. Cluster embeddings are computed by masking out image regions outside the bounding boxes associated with each cluster and then obtaining the CLIP embedding for the remaining regions. Cosine distances are calculated between the text embedding and each cluster embedding, and clusters with less than 85\% of the highest prompt-to-cluster similarity score are classified as outliers. Similarly, the authors of TFPOC~\cite{DBLP:conf/wacv/Shi0Z24} enabled textual inputs to define object classes. Their method begins with a coarse similarity map generated using CLIP-Surgery~\cite{DBLP:journals/corr/abs-2304-05653}, an enhanced variant of CLIP. From this map, reference objects are identified through a series of steps: the similarity map is binarized, the largest connected component (likely containing the target objects) is isolated, and bounding boxes are created around sub-regions within this component for further refinement. These bounding boxes are then used as prompts for Segment Anything Model~\cite{DBLP:conf/iccv/KirillovMRMRGXW23}, producing a high-quality similarity map. In another approach, the authors of~\cite{DBLP:conf/aaai/0018C24} employed CLIP to generate a prompt mask inspired by MaskCLIP~\cite{DBLP:conf/eccv/ZhouLD22}. Cosine distance was utilized to compute the matching score between predefined text features and the corresponding visual features. In PseCo~\cite{DBLP:conf/cvpr/HuangD0ZS24}, similar to the few-shot setting, the authors used the CLIP text embeddings as classification weights when only known class names. \csname rev4round3\endcsname{Finally, CountDiff~\cite{DBLP:conf/eccv/HuiWRL24} can be straightforwardly switched to perform open-world text-guided CAC by obtaining the object-specific embedding leveraging the pre-trained text encoder in the text-to-image diffusion model }

%%%%%%%%%%%%%%%%%%%%%%%%%%%%%%%%%%%%%%%%
% DATASETS
%%%%%%%%%%%%%%%%%%%%%%%%%%%%%%%%%%%%%%%%
\section{Datasets}
\label{sec:dataset}

%\subsection{Datasets}
%\label{sec:sec:datasets}
There is a lack of publicly available data for the CAC task.
Indeed, most existing counting datasets are designed for specific object categories. Notable examples include UCF-QNRF~\cite{DBLP:conf/eccv/IdreesTAZARS18}, ShanghaiTech~\cite{DBLP:conf/cvpr/ZhangZCGM16}, and WorldExpo'10~\cite{DBLP:conf/cvpr/ZhangLWY15} for crowd counting; TRANCOS~\cite{DBLP:conf/ibpria/Guerrero-Gomez-Olmedo15} and NDISPark~\cite{DBLP:conf/visapp/CiampiSCGA21, ciampi_ndispark_6560823} for vehicle counting; Pest Sticky Traps~\cite{DBLP:journals/ecoi/CiampiZICBFAC23, ciampi_2023_7801239} for pest abundance estimation; and VGG Cell~\cite{DBLP:journals/cmbbeiv/XieNZ18} for cell estimation. Meanwhile, datasets that include multiple object categories, such as the MSCOCO dataset~\cite{DBLP:conf/eccv/LinMBHPRDZ14}, are unsuitable for counting tasks because they were designed for object detection, typically featuring images with only a small number of object instances. In this section, we outline the limited CAC benchmarks available in the literature. Among them, FSC-147~\cite{DBLP:conf/cvpr/RanjanSNH21} is considered the gold standard. Additionally, although originally intended for vehicle counting, CARPK~\cite{DBLP:conf/iccv/HsiehLH17} is widely used to evaluate generalization capabilities. Fig.~\ref{fig:dataset} illustrates some samples from these two latter benchmarks.

%The gold standard benchmark for the class-agnostic counting task is FSC-147~\cite{DBLP:conf/cvpr/RanjanSNH21}, which features 147 object classes (see also Fig.~\ref{fig:dataset}). In contrast, most existing counting datasets are designed for specific object categories. Notable examples include UCF-QNRF~\cite{DBLP:conf/eccv/IdreesTAZARS18}, ShanghaiTech~\cite{DBLP:conf/cvpr/ZhangZCGM16}, and WorldExpo'10~\cite{DBLP:conf/cvpr/ZhangLWY15} for crowd counting; TRANCOS~\cite{DBLP:conf/ibpria/Guerrero-Gomez-Olmedo15}, NDISPark~\cite{DBLP:conf/visapp/CiampiSCGA21, ciampi_ndispark_6560823}, and CARPK~\cite{DBLP:conf/iccv/HsiehLH17} for vehicle counting; and VGG Cell~\cite{DBLP:journals/cmbbeiv/XieNZ18} for cell estimation. Meanwhile, datasets that include multiple object categories, such as the MSCOCO dataset~\cite{DBLP:conf/eccv/LinMBHPRDZ14}, are unsuitable for counting tasks because they were designed for object detection, typically featuring images with only a small number of object instances. In the following, we describe the main CAC benchmarks.

\paragraph{FSC-147}
The FSC-147 dataset comprises 6,135 images spanning 147 object categories, including plants, animals, vehicles, and food. %These images were sourced from Flickr, Google, and Bing using open-source image scrapers and then manually inspected and filtered. 
Annotations were manually created by marking dots over the approximate centroids of each object instance---as usual for the object counting task~\cite{DBLP:conf/nips/LempitskyZ10}. %Occluded instances were annotated only if at least 10\% of the object was visible. 
Additionally, the dataset includes the coordinates of three bounding boxes per image, representing exemplar instances. Each image contains exemplars belonging to a single category, whose natural language name is stored in a text file. If multiple categories were present in an image, one was chosen arbitrarily. Another characteristic of FSC-147 is the significant variation in object counts per image, ranging from 7 to 3,731, with an average of 56 objects per image. The dataset is divided into three splits: 89 object categories are allocated to the training set, while 29 categories are assigned to the validation and testing sets. These splits consist of 3,659, 1,286, and 1,190 images, respectively.

\paragraph{FSC-133}
A revised version of FSC-147, called FSC-133, was introduced in~\cite{DBLP:journals/corr/abs-2205-10203}. The authors identified and corrected several errors in FSC-147. \csname rev1\endcsname{Specifically, they found that (i) 11 training images also appeared in validation or test sets, (ii) some images were near-duplicates with pixel-wise differences close to zero, and (iii) certain duplicates had inconsistent ground truth counts, with discrepancies up to 25\%.} To address these issues, FSC-133 merged very similar categories and retained only the images with the most accurate counts. The revised dataset consists of 5,898 images across 133 categories, with splits containing 3,877 images for training, 954 for validation, and 1,067 for testing. However, despite the improvements introduced with FSC-133, subsequent works have largely continued to use FSC-147, which remains the standard for this task.

\paragraph{FSC-147-D and REC-8K}
Another enhanced version of FSC-147, called FSC-147-D, was introduced by~\cite{AminiNaieni23} to better support the open-world text-guided object counting paradigm. In this version, Amini-Naieni et al. replaced the simpler object descriptions with more fine-grained and structured natural language sentences for each image, specifying the objects to be counted with greater precision. In a similar direction, REC-8K~\cite{DBLP:conf/cvpr/DaiLC24} is a collection of approximately 8,000 images drawn from existing datasets, including FSC-147, where the textual descriptions of object categories are enriched with attributes.

\paragraph{OmniCount-191 and MCAC}
Moreover, two datasets have addressed a key limitation of FSC-147---its limited number of images featuring multiple object categories---by including images with multiple categories: OmniCount-191~\cite{DBLP:conf/aaai/MondalNZ025} and MCAC~\cite{DBLP:conf/eccv/HobleyP24}. However, both present significant drawbacks. OmniCount-191 contains a low number of objects per image, and many images within the same category are visually similar---often being augmented versions of the same source---even across training and test splits. MCAC, on the other hand, is synthetic and lacks appropriate annotations for open-world text-guided approaches, limiting its applicability to realistic scenarios. \csname rev4round3\endcsname{Notably, still within the context of synthetic data,~\cite{DBLP:conf/eccv/DAlessandroMH24} and~\cite{DBLP:conf/wacv/DoubinskyACB24} proposed approaches relying on text-to-image latent diffusion models to produce counting data across diverse object categories.}

\paragraph{CARPK}
Finally, several existing papers have also evaluated their performance on the CARPK dataset~\cite{DBLP:conf/iccv/HsiehLH17}, a collection of drone-view images of parking lots specifically designed for vehicle counting. \csname rev3\endcsname{This benchmark was first utilized for CAC by the work that introduced FSC-147 and has since been used by subsequent works, primarily to assess cross-dataset generalization on the specific \textit{vehicle} category.} The dataset comprises 1,448 images, divided into a training set of 989 images and a test set of 459 images, with approximately 90,000 annotated vehicles in total. \csname rev3\endcsname{In CAC, this dataset is used either without any annotations or with a set of 12 randomly selected exemplars.}
%The annotations feature bounding boxes, and density maps are generated using dots placed at the centers of these bounding boxes. Additionally, some of the bounding boxes are utilized to identify exemplars.

%Check also the FSCD-LVIS dataset \cite{DBLP:conf/eccv/NguyenPNH22}; some works reported results for this (even if just a few works). Maybe we also have to consider this dataset and eventually report results, computing the ones that are not present in the literature?

% Consider also the new synthetic dataset

\begin{table*}[!htbp]
    \caption{\textbf{Results on the FSC-147 dataset~\cite{DBLP:conf/cvpr/RanjanSNH21}.} We report the performance in terms of MAE and RMSE reached by the methodologies discussed in this survey over the gold standard CAC benchmark---FSC-147. The best results are in bold.}
    \label{tab:fsc_results}
    \setlength{\tabcolsep}{2pt}
    %\small
    \footnotesize
    \newcolumntype{C}{>{\centering\arraybackslash}X}
    \center\begin{tabularx}{0.65\linewidth}{lCCCC}
            \toprule
            & \multicolumn{2}{c}{Val Set} & \multicolumn{2}{c}{Test Set} \\
            \cmidrule(lr){2-3} \cmidrule(lr){4-5} 
            Method & \textit{MAE} $\downarrow$ & \textit{RMSE} $\downarrow$ & \textit{MAE} $\downarrow$ & \textit{RMSE} $\downarrow$ \\
            \midrule
            \midrule
            \textit{Baseline} & & & & \\
            Mean & 53.38 & 124.53 & 47.55 & 147.67 \\
            Median & 48.68 & 129.70 & 47.73 & 152.46 \\
            \midrule
            \textit{Reference-based (3 exemplars)} & & & & \\
            %GMN~\cite{DBLP:conf/accv/LuXZ18} & 29.66 & 89.81 & 26.52 & 124.57 \\
            FamNet~\cite{DBLP:conf/cvpr/RanjanSNH21} & 23.75 & 69.07 & 22.08 & 99.54 \\
            %FamNet$^*$~\cite{DBLP:conf/cvpr/RanjanSNH21} & 24.32 & 70.94 & 22.56 & 101.54 \\
            CFOCNet$^{*}$~\cite{DBLP:conf/wacv/YangSHC21} & 21.19 & 61.41 & 22.10 & 112.71 \\
            %CFOCNet$^\$$~\cite{DBLP:conf/wacv/YangSHC21} & 27.82 & 71.99 & 28.60 & 123.96 \\
            VCN~\cite{DBLP:conf/cvpr/RanjanH22} & 19.38 & 60.15 & 18.17 & 95.60 \\
            %VCN$^*$~\cite{DBLP:conf/cvpr/RanjanH22} & 20.24 & 62.99 & 19.30 & 99.08 \\
            RCAC~\cite{DBLP:conf/eccv/GongZ0DS22} & 20.54 & 60.78 & 20.21 & 81.86 \\
            BMNet+~\cite{DBLP:conf/cvpr/Shi0FL022} & 15.74 & 58.53 & 14.62 & 91.83 \\
            SPDCN~\cite{DBLP:conf/bmvc/LinYM0LLHYC22} & 14.59 & 49.97 & 13.51 & 96.80 \\
            MACnet~\cite{DBLP:journals/prl/McCarthyVOTAH23} & 20.65 & - & 18.38 & - \\
            SAFECount~\cite{DBLP:conf/wacv/YouYLLCL23} & 15.28 & 47.20 & 14.32 & 85.54 \\
            LOCA~\cite{DBLP:conf/iccv/EukicLZK23} & 10.24 & 32.56 & 10.79 & 56.97 \\
            %MACnet$^*$~\cite{DBLP:journals/prl/McCarthyVOTAH23} & 20.81 & - & 18.41 & - \\
            ConCoNet (FamNet)~\cite{DBLP:journals/prl/SolivenVOTAH23} & 18.77 & 58.96 & 18.02 & 91.63 \\
            ConCoNet (BMNet)~\cite{DBLP:journals/prl/SolivenVOTAH23} & 15.03 & 57.94 & 14.21 & 91.12 \\ 
            %ASFNet~\cite{10145320} & 20.06 & 64.85 & 18.63 & 104.37 \\
            %SAM~\cite{DBLP:journals/corr/abs-2304-10817} & 31.20 & 100.83 & 27.97 & 131.24 \\
            SSD~\cite{DBLP:conf/ijcai/Xu0Z24} & 9.73 & 29.72 & 9.58 & 64.13 \\
            SATCount~\cite{DBLP:journals/nn/WangYWLC24} & 12.27 & 45.91 & 10.77 & 61.14 \\
            CSTrans~\cite{DBLP:journals/pr/GaoH24} & 18.10 & 58.45 & 16.38 & 93.51 \\
            CountVers~\cite{DBLP:journals/kbs/YangCDWZ24} & 11.45 & 42.29 & 11.28 & 90.10 \\
            %TFPOC~\cite{DBLP:conf/wacv/Shi0Z24}$^{\$\$}$ & - & - & 19.95 & 132.16 \\
            %SAM$^{\$\$}$~\cite{DBLP:conf/wacv/Shi0Z24} & - & - & 42.48 & 137.50 \\
            CACViT~\cite{DBLP:conf/aaai/WangX0024} & 10.63 & 37.95 & 9.13 & 48.96 \\
            DAVE~\cite{DBLP:conf/cvpr/PelhanLZK24} & 8.91 & \textbf{28.08} & \textbf{8.66} & \textbf{32.36} \\
            TFPOC~\cite{DBLP:conf/wacv/Shi0Z24} & - & - & 19.95 & 132.16 \\
            UPC~\cite{DBLP:conf/aaai/0018C24} & 16.87 & 59.45 & 16.68 & 105.08 \\
            %T-Rex2~\cite{DBLP:journals/corr/abs-2403-14610} & - & - & 10.94 & 43.35 \\
            PseCo~\cite{DBLP:conf/cvpr/HuangD0ZS24} & 15.31 & 68.34 & 13.05 & 112.86 \\
            \textcolor{black}{CountDiff~\cite{DBLP:conf/eccv/HuiWRL24}} & \textcolor{black}{\textbf{8.43}} & \textcolor{black}{31.03} & \textcolor{black}{9.24} & \textcolor{black}{53.41} \\
            %\textcolor{darkgreen}{GeCo~\cite{DBLP:conf/nips/PelhanLZK24}} & \textcolor{darkgreen}{9.52} & \textcolor{darkgreen}{43.00} & \textcolor{darkgreen}{\textbf{7.91}} & \textcolor{darkgreen}{54.28} \\
            %CFENet~\cite{DBLP:journals/ivc/ZhangZCWH25} & 11.43 & 37.42 & 9.38 & 33.37 \\
            % CAP~\cite{10671540} & 8.78 & 28.42 & 9.07 & 51.19 \\ 
            %MEAMNet~\cite{DBLP:journals/prl/ZhangHZLCL24} & 12.59 & 40.95 & 13.48 & 84.11 \\
            %CountGD~\cite{DBLP:journals/corr/abs-2407-04619} & 7.46 & 29.54 & 8.31 & 91.05 \\
            %\noalign{\vskip 2pt} 
            %\hdashline
            %\noalign{\vskip 2pt}
            CounTR$^{\$}$~\cite{DBLP:conf/bmvc/LiuZZX22} & 13.13 & 49.83 & 11.95 & 91.23 \\
            GCNet$^{\$}$~\cite{DBLP:journals/pr/WangLZTG24} & 19.61 & 66.22 & 17.86 & 106.98 \\
            \midrule
            \textit{Reference-less} & & & & \\
            RepRPN-Counter (Top5)~\cite{DBLP:conf/accv/RanjanN22} & 29.24 & 98.11 & 26.66 & 129.11 \\
            %RepRPN-Counter (Top3)~\cite{DBLP:conf/accv/RanjanN22} & 30.40 & 98.73 & 27.45 & 129.69 \\
            %RepRPN-Counter (Top1)~\cite{DBLP:conf/accv/RanjanN22} & 31.69 & 100.31 & 28.32 & 128.76 \\
            RCC~\cite{DBLP:journals/corr/abs-2205-10203} & 17.49 & 58.81 & 17.12 & 104.53 \\
            CounTR~\cite{DBLP:conf/bmvc/LiuZZX22} & 17.40 & 70.33 & \textbf{14.12} & 108.01 \\
            GCNet~\cite{DBLP:journals/pr/WangLZTG24} & 19.50 & 63.13 & 17.83 & \textbf{102.89} \\
            %\noalign{\vskip 2pt} 
            %\hdashline
            %\noalign{\vskip 2pt}
            FamNet (Top5)$^{\$\$}$$^{\dagger}$~\cite{DBLP:conf/cvpr/RanjanSNH21} & 32.15 & 98.75 & 32.27 & 131.46 \\
            LOCA$^{\$\$}$~\cite{DBLP:conf/iccv/EukicLZK23} & 17.43 & 54.96 & 16.22 & 103.96 \\
            DAVE$^{\$\$}$~\cite{DBLP:conf/cvpr/PelhanLZK24} & \textbf{15.54} & \textbf{52.67} & 15.14 & 103.49 \\
            % CAP$^{\$\$}$~\cite{10671540} & 16.92 & 53.04 & 13.84 & 99.92 \\ 
            %\textcolor{darkgreen}{GeCo$^{\$\$}$~\cite{DBLP:conf/nips/PelhanLZK24}} & \textcolor{darkgreen}{\textbf{14.81}} & \textcolor{darkgreen}{64.95} & \textcolor{darkgreen}{\textbf{13.30}} & \textcolor{darkgreen}{108.72} \\
            \midrule
            \textit{Open-world Text-guided} & & & & \\
            ZSC~\cite{DBLP:conf/cvpr/XuL0RS23} & 26.93 & 88.63 & 22.09 & 115.17 \\
            CounTX$^{**}$~\cite{AminiNaieni23} & 17.70 & 63.61 & 15.73 & 106.88 \\
            CLIP-Count~\cite{DBLP:conf/mm/JiangLC23} & 18.79 & 61.18 & 17.78  & 106.62 \\
            VLCounter~\cite{DBLP:conf/aaai/KangMKH24} & 18.06 & 65.13 & 17.05 & 106.16 \\
            GroundingREC$^{++}$~\cite{DBLP:conf/cvpr/DaiLC24} & \textbf{10.06} & 58.62 & \textbf{10.12} & 107.19 \\
            VA-Count~\cite{DBLP:conf/eccv/ZhuYYGWZH24} & 17.87 & 73.22 & 17.88 & 129.31 \\
            %\noalign{\vskip 2pt} 
            %\hdashline
            %\noalign{\vskip 2pt}
            %\textcolor{darkgreen}{CountGD~\cite{DBLP:conf/nips/Amini-NaieniHZ24}} & \textcolor{darkgreen}{12.14} & \textcolor{darkgreen}{\textbf{47.51}} & \textcolor{darkgreen}{12.98} & \textcolor{darkgreen}{\textbf{98.35}} \\
            DAVE$^{\ddagger}$~\cite{DBLP:conf/cvpr/PelhanLZK24} & 15.48 & \textbf{52.57} & 14.90 & 103.42 \\
            TFPOC$^{\ddagger}$$^+$~\cite{DBLP:conf/wacv/Shi0Z24} & 47.21 & 127 & 24.79 & 137.15 \\
            UPC~\cite{DBLP:conf/aaai/0018C24}$^{\ddagger}$ & 16.92 & 58.92 & 16.81 & 105.83 \\
            PseCo~\cite{DBLP:conf/cvpr/HuangD0ZS24}$^{\ddagger}$ & 23.90 & 100.33 & 16.58 & 129.77 \\
            \textcolor{black}{CountDiff~\cite{DBLP:conf/eccv/HuiWRL24}$^{\ddagger}$} & \textcolor{black}{15.50} & \textcolor{black}{54.33} & \textcolor{black}{14.83} & \textcolor{black}{\textbf{103.15}} \\
            % CountGD~\cite{DBLP:journals/corr/abs-2407-04619} & 12.14 & 47.51 & 12.98 & 98.35 \\
            \bottomrule \\
    \end{tabularx}
    \par\vspace{-5pt}
    {\setlength{\baselineskip}{0.7\baselineskip}\footnotesize
    %* without test-time adaptation; 
    * re-implemented by~\cite{DBLP:conf/cvpr/Shi0FL022}; 
    %\$ re-implemented by~\cite{DBLP:conf/eccv/GongZ0DS22}; 
    \$ reference-based modification of a reference-less method;
    \$\$ reference-less modification of a reference-based method;
    $\dagger$ implemented by~\cite{DBLP:conf/accv/RanjanN22};
    $\ddagger$ prompt-based modification of a reference-based method;
    $+$ results on the validation set computed by~\cite{DBLP:conf/cvpr/DaiLC24};
    ** trained on the FSC-147-D dataset~\cite{AminiNaieni23}; 
    ++ trained on the REC-8K dataset~\cite{DBLP:conf/cvpr/DaiLC24}}.
\end{table*}

%%%%%%%%%%%%%%%%%%%%%%%%%%%%%%%%%%%%%%%%
% RESULTS
%%%%%%%%%%%%%%%%%%%%%%%%%%%%%%%%%%%%%%%%
\section{Results and Discussion}
\label{sec:results} 
In this section, we present the results of the evaluated methods, following the same taxonomy of previous sections---reference-based, reference-less, and open-world text-guided approaches. First, we describe the experimental setting and then provide a critical analysis of the outcomes across various points of discussion.

\subsection{Experimental Setting}
Some techniques have been adapted to carry out more than one of these tasks (see also Sec.~\ref{sec:existing_works}); in these cases, we report the results for all of them. We considered the two most commonly used benchmarks: FSC-147~\cite{DBLP:conf/cvpr/RanjanSNH21} and CARPK~\cite{DBLP:conf/iccv/HsiehLH17}. For FSC-147, in line with existing literature, we report results for both the validation and test subsets. Differently, the CARPK dataset is used to evaluate the generalization capabilities of the models in counting instances of a specific object class---vehicles. Previous works have explored two different settings: i) training models on FSC-147 without using any data from CARPK nor the car category in FSC-147, and ii) fine-tuning models using training data from CARPK, where a set of 12 exemplars is randomly sampled from the training set and used as exemplars across all training and test images.

We show the outcomes in terms of two popular counting metrics, following previous literature, i.e., using the mean absolute error (MAE) and the root mean squared error (RMSE), defined as $\text{MAE} = \frac{1}{N_T} \sum_{n=1}^{N_T} \left| \tilde{c}_n - c_n \right|$ and $\text{RMSE} = \sqrt{\frac{1}{N_T} \sum_{n=1}^{N_T} ( \tilde{c}_n - c_n )^2}$, where $N_T$ is the number of test images, and $\tilde{c}_n$ and $c_n$ are the ground truth and predicted counts, respectively.

\subsection{Reference-based Results}
\label{sec:sec:reference-based-results}
Table~\ref{tab:fsc_results} shows the results achieved by the surveyed reference-based methods on FSC-147. We also report the results concerning two trivial baselines, as in~\cite{DBLP:conf/cvpr/RanjanSNH21}, i.e., always output the average object count for training images and always output the median count for training images. 

The pioneering FamNet~\cite{DBLP:conf/cvpr/RanjanSNH21} laid the groundwork for this paradigm with an MAE and an RMSE of 22.08 and 99.54 on the test subset. Subsequent works progressively improve the performance, sometimes revising and adapting this foundation model. For instance, VCN~\cite{DBLP:conf/cvpr/RanjanH22}, RCAC~\cite{DBLP:conf/eccv/GongZ0DS22}, and SPDCN~\cite{DBLP:conf/bmvc/LinYM0LLHYC22} focused on the robustness of exemplar feature representations, improving generalization, and highlighting the importance of robust training data diversity. Especially, VCN and SPDCN reached an impressive MAE of 18.17 and 13.51, respectively. Differently, MACnet introduced segmentation masks as an alternative to bounding boxes for exemplar representation, demonstrating that finer object delineation can improve model robustness. The subsequent method that has marked a breakthrough is BMNet+, which was one of the first works that improved the matching technique between image and exemplar features, moving away from the naive idea of the fixed inner product between them. SAFECount~\cite{DBLP:conf/wacv/YouYLLCL23} also followed the direction of improving the image-exemplars matching mechanism and achieved an MAE and RMSE of 14.32 and 85.54, respectively (again, on the test subset). 
Other methods that stood out are SSD~\cite{DBLP:conf/ijcai/Xu0Z24}, LOCA~\cite{DBLP:conf/iccv/EukicLZK23}, and CACViT~\cite{DBLP:conf/aaai/WangX0024}, which sensibly lowered both MAE and RMSE. The latter approach proposed a particularly interesting "extract-and-match" paradigm, which basically totally relies on the attention mechanism of Video Transformer~\cite{DBLP:conf/iclr/DosovitskiyB0WZ21}. Additionally, ConCoNet is worth mentioning since it proposed a training strategy potentially applicable to all the other methods that brought the idea of leveraging negative exemplars to better distinguish between target and background regions. \csname rev3\endcsname{However, the model achieving state-of-the-art results is DAVE~\cite{DBLP:conf/cvpr/PelhanLZK24}. By employing a detect-and-verify strategy, DAVE combines bounding-box detection with clustering-based verification to eliminate outliers effectively. On the leaderboard, DAVE demonstrates the lowest MAE and RMSE values to date (8.66 and 32.36, respectively, on the test subset), showcasing its dominance in the reference-based paradigm. This performance leap is not solely due to architectural complexity but reflects a shift in design philosophy. Earlier regression-based methods like FamNet are vulnerable to background clutter, which can produce false density peaks and catastrophic counting errors. DAVE addresses this by decoupling the process: first generating candidate detections, then verifying them against exemplar clusters. This verification stage acts as a semantic filter, rejecting outliers and enforcing consistency with the exemplar distribution.} \csname rev4round3\endcsname{Finally, it is worth noting that the best MAE on the validation subset is achieved by CountDiff~\cite{DBLP:conf/eccv/HuiWRL24}. However, this result relies on a test-time adaptation strategy, making the comparison not entirely fair since additional training occurs during inference. Without this component, performance is less competitive---MAE/RMSE of 9.15/31.97 on the validation subset and 9.97/54.89 on the test subset.}

\subsection{Reference-less Results}
\label{sec:sec:reference-less-results}
In Tab.~\ref{tab:fsc_results}, we report the results obtained by the surveyed reference-less approaches on FSC-147. The first reference-less approach, RepRPN-Counter~\cite{DBLP:conf/accv/RanjanN22}, introduced the concept of repetitive region proposals and reached an MAE of 26.66 and an RMSE of 129.11 on the test subset. These error metrics are significantly higher than those obtained with reference-based approaches, as expected, since reference-less techniques do not utilize bounding box exemplar annotations. Subsequent methods have significantly improved performance. CounTR~\cite{DBLP:conf/bmvc/LiuZZX22}, a transformer-based reference-less framework, leverages the self-similarity of image regions and achieves competitive results, with an MAE of 14.12 and an RMSE of 108.01 on the test subset. Another notable contribution is RCC~\cite{DBLP:journals/corr/abs-2205-10203}. Besides achieving competitive results (an MAE of 17.12 and an RMSE of 104.53), it employs a weakly supervised learning approach that does not need point-level annotations, thereby representing a particularly interesting approach in data scarcity scenarios. In the same direction, also GCNet~\cite{DBLP:journals/pr/WangLZTG24} does not use dot labels and obtained comparable results to RCC. Overall, there is no clear winner in this leaderboard. Indeed, on the one hand, the reference-less version of DAVE~\cite{DBLP:conf/cvpr/PelhanLZK24} achieves the best results in the validation split with an MAE and RMSE of 15.54 and 52.67, respectively; on the other hand, CounTR and GCNet shine in the test subset with an MAE of 14.12 and an RMSE of 102.89. 

\csname rev3\endcsname{These results highlight a key trade-off in reference-less CAC: while these methods eliminate the need for exemplar annotations and offer greater flexibility, they often suffer from semantic ambiguity and background clutter. Without explicit guidance, they rely on visual repetition to infer the dominant object class, which can lead to false positives in complex scenes.}

\subsection{Open-world Text-guided Results}
\label{sec:sec:zero-shot-results}
In the last part of Tab.~\ref{tab:fsc_results}, we show the results concerning the considered open-world text-guided approaches, which represent a recent and promising advancement in class-agnostic counting, driven by the progress of large multi-modal models. However, similar to the reference-less case, this flexibility comes at the expense of lower performance compared to reference-based techniques.
The first milestone, ZSC~\cite{DBLP:conf/cvpr/XuL0RS23}, obtained an MAE and an RMSE of 22.09 and 115.17 on the test subset. Among the subsequent methods that steadily improved performance, CounTX~\cite{AminiNaieni23} emerges as a particularly competitive approach, in some way adapting its counterpart CounTR~\cite{DBLP:conf/bmvc/LiuZZX22} to this new setting. It achieves an impressive MAE of 15.73 and an RMSE of 106.88 on the test subset. \csname rev4round3\endcsname{However, the best results were contended by two open-world text-guided variants of reference-based methods---DAVE~\cite{DBLP:conf/cvpr/PelhanLZK24} and CountDiff~\cite{DBLP:conf/eccv/HuiWRL24}---and GroundingREC~\cite{DBLP:conf/cvpr/DaiLC24}. The first two excel in RMSE, indicating greater robustness to outliers, while GroundingREC achieves the best MAE (10.06 and 10.12 on the validation and test subsets, respectively).}

\csname rev3round2\endcsname{Beyond raw performance dictated by architectural choices, the effectiveness of text-guided counting is also influenced by the richness of the textual prompts used. While most methods rely on simple class names, recent approaches have explored more structured and semantically expressive prompts. For instance, CounTX is trained on the FSC-147-D dataset, which replaces basic labels with more descriptive textual sentences. Similarly, GroundingREC is trained on the REC-8K dataset, where prompts include object attributes. Notably, both CounTX and GroundingREC rank among the top-performing models in our evaluation, suggesting that richer prompts can lead to more accurate and robust counting. These results highlight the potential of moving beyond minimal textual input toward more informative and context-aware prompting. Future research could explore automatic prompt generation, adaptive refinement mechanisms, and multimodal reasoning strategies to further improve semantic alignment and generalization in open-world scenarios, possibly also considering the contextual capabilities of large multimodal models as part of broader reasoning pipelines.}

\subsection{\textcolor{black}{Trade-offs Between Guidance and Accuracy}}
\csname rev3\endcsname{A central theme emerging from these results is the fundamental trade-off between performance and convenience across the three CAC paradigms. While reference-based methods consistently achieve the best results (e.g., DAVE~\cite{DBLP:conf/cvpr/PelhanLZK24} achieves an MAE of 8.66 on FSC-147 test set), they require human-annotated exemplars at inference time, limiting scalability and automation. In contrast, reference-less and open-world text-guided approaches offer greater convenience---eliminating or automating the need for exemplars—but at a measurable cost in accuracy.
This ``price of convenience'' is starkly illustrated by the more than 70\% increase in MAE observed when moving from DAVE (8.66) to its reference-less variant (15.14) or its open-world text-guided version (14.90).}

\csname rev3\endcsname{However, beyond the illustrative case of DAVE, all reference-less and open-world text-guided approaches consistently show lower performance compared to reference-based techniques. Notably, the RMSE is particularly affected by the change of paradigm: the best results on the FSC-147 test set are achieved by DAVE with an RMSE of 32.36, GCNet~\cite{DBLP:journals/pr/WangLZTG24} with 102.89, and the text-guided version of DAVE with 103.42, representing the top-performing models in the reference-based, reference-less, and text-guided paradigms, respectively. We argue that this large error gap in RMSE can often be attributed to the use of poor-quality or even incorrect exemplars---a challenge that becomes more pronounced in the absence of reference bounding boxes---and may result in catastrophic errors on some images.
The logical conclusion is that, as of today, the core challenge of effective guidance in class-agnostic counting entails a quantifiable cost in accuracy---a trade-off that should be central to both the evaluation and future development of CAC methods.}

\begin{table}[!t]
    \caption{\textbf{Results on the CARPK dataset~\cite{DBLP:conf/iccv/HsiehLH17}.} We report the performance in terms of MAE and RMSE obtained with some of the methodologies discussed in this survey over the CARPK benchmark. Following previous literature, CARPK is exploited for testing cross-dataset generalization over a specific object class---vehicles. The best results are in bold. }
    \label{tab:carpk_results}
    \setlength{\tabcolsep}{2pt}
    \footnotesize
    \newcolumntype{C}{>{\centering\arraybackslash}m{.19\linewidth}}
    \newcolumntype{Z}{>{\centering\arraybackslash}m{.14\linewidth}}
    \center
    \begin{tabularx}{0.99\linewidth}{lCZZ}
            \toprule
            Method & Fine-Tuned & \textit{MAE} $\downarrow$ & \textit{RMSE} $\downarrow$ \\
            \midrule
            \midrule
            \textit{Baseline} & & & \\
            Mean & N/A & 65.63 & 72.26 \\
            Median & N/A & 67.88 & 74.58 \\
            \midrule
            \textit{Reference-based (12 exemplars)} & & & \\
            %GMN$^*$~\cite{DBLP:conf/accv/LuXZ18} & \cmark & 7.48 & 9.90 \\
            FamNet~\cite{DBLP:conf/cvpr/RanjanSNH21} & \cmark & 18.19 & 33.66 \\
            % CFOCNet~\cite{DBLP:conf/wacv/YangSHC21} & - & - & - \\
            % VCN~\cite{DBLP:conf/cvpr/RanjanH22} & - & - & - \\
            RCAC~\cite{DBLP:conf/eccv/GongZ0DS22} & \cmark & 13.62 & 19.08 \\
            % MACnet~\cite{DBLP:journals/prl/McCarthyVOTAH23} & - & - & - \\
            BMNet+~\cite{DBLP:conf/cvpr/Shi0FL022} & \cmark & 5.76 & 7.83 \\
            SPDCN~\cite{DBLP:conf/bmvc/LinYM0LLHYC22} & \cmark & 10.07 & 14.12 \\
            SAFECount~\cite{DBLP:conf/wacv/YouYLLCL23} & \cmark & 5.33 & 7.04 \\
            CSTrans~\cite{DBLP:journals/pr/GaoH24} & \cmark & 5.06 & 6.53 \\
            CountVers~\cite{DBLP:journals/kbs/YangCDWZ24} & \cmark & 5.19 & 6.55 \\
            CACViT~\cite{DBLP:conf/aaai/WangX0024} & \cmark & \textbf{4.91} & \textbf{6.49} \\
            %CFENet~\cite{DBLP:journals/ivc/ZhangZCWH25} & \cmark & \textbf{4.56} & \textbf{5.74} \\
            %MEAMNet~\cite{DBLP:journals/prl/ZhangHZLCL24} & \cmark & 4.93 & 6.50 \\
            \noalign{\vskip 2pt} 
            \hdashline
            \noalign{\vskip 2pt}
            FamNet~\cite{DBLP:conf/cvpr/RanjanSNH21} & \xmark & 28.84 & 44.47 \\
            RCAC~\cite{DBLP:conf/eccv/GongZ0DS22} & \xmark & 17.98 & 24.21 \\
            BMNet+~\cite{DBLP:conf/cvpr/Shi0FL022} & \xmark & 10.44 & 13.77 \\
            SPDCN~\cite{DBLP:conf/bmvc/LinYM0LLHYC22} & \xmark & 18.15 & 21.61 \\
            ConCoNet (FamNet)~\cite{DBLP:journals/prl/SolivenVOTAH23} & \xmark & 23.55 & 32.41 \\
            ConCoNet (BMNet)~\cite{DBLP:journals/prl/SolivenVOTAH23} & \xmark & 16.40 & 21.54 \\
            %ASFNet~\cite{10145320} & - & - & - \\
            SAFECount~\cite{DBLP:conf/wacv/YouYLLCL23} & \xmark & 16.66 & 24.08 \\
            LOCA~\cite{DBLP:conf/iccv/EukicLZK23} & \xmark & 9.97 & 12.51 \\
            SATCount~\cite{DBLP:journals/nn/WangYWLC24} & \xmark & 9.51 & 12.67 \\
            CACViT~\cite{DBLP:conf/aaai/WangX0024} & \xmark & 8.30 & 11.18 \\
            SSD~\cite{DBLP:conf/ijcai/Xu0Z24} & \xmark & 9.58 & 12.15 \\
            % DAVE~\cite{DBLP:conf/cvpr/PelhanLZK24} & - & - & - \\ 
            TFPOC~\cite{DBLP:conf/wacv/Shi0Z24} & \xmark & 10.97 & 14.24 \\
            UPC~\cite{DBLP:conf/aaai/0018C24} & \xmark & \textbf{7.83} & \textbf{9.74} \\
            \textcolor{black}{CountDiff~\cite{DBLP:conf/eccv/HuiWRL24}} & \textcolor{black}{\xmark} & \textcolor{black}{8.36} & \textcolor{black}{10.84} \\
            %CFENet~\cite{DBLP:journals/ivc/ZhangZCWH25} & \xmark & 10.68 & 13.25 \\
            % CAP~\cite{10671540} & \xmark & 9.11 & 11.02 \\
            %\midrule
            %\textit{Reference-less} & & & \\
            % RepRPN-Counter~\cite{DBLP:conf/accv/RanjanN22} & - & - & - \\
            %RCC~\cite{DBLP:journals/corr/abs-2205-10203} & \cmark & 9.21 & 11.33 \\
            %RCC~\cite{DBLP:journals/corr/abs-2205-10203} & \xmark & 12.31 & 15.40 \\
            %CounTR~\cite{DBLP:conf/bmvc/LiuZZX22} & \cmark & 5.75 & 7.45 \\
            % GCNet~\cite{DBLP:journals/pr/WangLZTG24} & - & - & - \\ 
            \midrule
            \textit{Open-world Text-guided} & & & \\
            % ZSC~\cite{DBLP:conf/cvpr/XuL0RS23} & - & - & - \\
            %CounTX~\cite{AminiNaieni23} & \cmark & 8.13 & 10.87 \\
            CounTX~\cite{AminiNaieni23} & \xmark & 11.64 & 14.85 \\ 
            CLIP-Count~\cite{DBLP:conf/mm/JiangLC23} & \xmark & 11.96 & 16.61 \\
            VLCounter~\cite{DBLP:conf/aaai/KangMKH24} & \xmark & \textbf{6.46} & \textbf{8.68} \\
            VA-Count~\cite{DBLP:conf/eccv/ZhuYYGWZH24} & \xmark & 10.63 & 13.20 \\
            %\textcolor{darkgreen}{CountGD~\cite{DBLP:conf/nips/Amini-NaieniHZ24}} & \textcolor{darkgreen}{\xmark} & \textcolor{darkgreen}{\textbf{3.83}} & \textcolor{darkgreen}{\textbf{5.41}} \\
            TFPOC$^{\ddagger}$~\cite{DBLP:conf/wacv/Shi0Z24} & \xmark & 11.01 & 14.34 \\
            UPC~\cite{DBLP:conf/aaai/0018C24}$^{\ddagger}$ & \xmark & 7.62 & 9.71 \\
            \textcolor{black}{CountDiff~\cite{DBLP:conf/eccv/HuiWRL24}$^{\ddagger}$} & \textcolor{black}{\xmark} & \textcolor{black}{10.32} & \textcolor{black}{12.92} \\
            % CountGD~\cite{DBLP:journals/corr/abs-2407-04619} & \xmark & 3.68 & 5.17 \\
            \bottomrule \\
    \end{tabularx}
    \par\vspace{-5pt} 
     {\setlength{\baselineskip}{0.7\baselineskip}\footnotesize
    %* GMN uses extra training data from the ILSVRC video dataset~\cite{DBLP:journals/ijcv/RussakovskyDSKS15}, which consists of video sequences of cars; 
    $\ddagger$ prompt-based modification of a reference-based method.}
\end{table}

\subsection{Cross-dataset Generalization}
\csname rev3\endcsname{Table~\ref{tab:carpk_results} summarizes the results on the CARPK dataset~\cite{DBLP:conf/iccv/HsiehLH17} among methods that reported outcomes on this benchmark.}

\csname rev3\endcsname{Fine-tuned reference-based approaches demonstrate substantial performance gains compared to non-fine-tuned ones, as expected, since the latter do not see the vehicle category at all. In contrast, experiments on CARPK with fine-tuned models diverge slightly from the CAC paradigm, as the model is tested on the same class seen during fine-tuning. Rather than evaluating generalization to unseen categories, as in the standard CAC setting, this scenario assesses the ability to learn from a small number of exemplars and generalize across a dataset containing objects of the same category.
Specifically, CACViT~\cite{DBLP:conf/aaai/WangX0024} leads the pack of fine-tuned techniques (MAE: 4.91, RMSE: 6.49). Its performance likely stems from the use of a pre-trained Vision Transformer that performs unified feature extraction and matching via attention mechanisms. The model processes the query image and exemplar patches as token sequences---referred to as query and exemplar tokens---which are concatenated and jointly fed into the transformer encoder. Within this unified token space, self-attention extracts contextual features from both inputs and simultaneously performs similarity matching between them. This design appears particularly effective in learning from very few exemplars (12 in this case), and demonstrates robustness in generalizing across the dataset.
BMNet+~\cite{DBLP:conf/cvpr/Shi0FL022} and SAFECount~\cite{DBLP:conf/wacv/YouYLLCL23} also deliver strong results (MAEs of 5.76 and 5.33, respectively), leveraging advanced similarity modules and data-driven refinement strategies that enhance adaptability.}

\csname rev3\endcsname{Non-fine-tuned methods generally show higher error rates, yet recent models like LOCA~\cite{DBLP:conf/iccv/EukicLZK23}, SATCount~\cite{DBLP:journals/nn/WangYWLC24}, CACViT~\cite{DBLP:conf/aaai/WangX0024}, and SSD~\cite{DBLP:conf/ijcai/Xu0Z24} narrow the gap with fine-tuned approaches. This set of experiments fully reflects the CAC setting, since models have not seen the vehicle category during training. The goal here is to evaluate their ability to count a specific object category in a new dataset under domain shift.
Notably, UPC~\cite{DBLP:conf/aaai/0018C24} achieves the best performance among non-fine-tuned methods (MAE: 7.84, RMSE: 9.74), and remains competitive even in open-world text-guided settings (MAE: 7.62, RMSE: 9.71). UPC’s strong generalization may be attributed to its iterative refinement mechanism, which progressively updates the predicted density map by reusing it as a new prompt. This feedback loop allows the model to converge toward consistent outputs and may help suppress spurious activations in unfamiliar domains. Additionally, the use of a contrastive training strategy---where the model learns to distinguish between prompts that match the target object and those that do not---encourages the network to focus on semantically meaningful regions. These mechanisms likely improve the model’s robustness to domain shift by reinforcing object-specific consistency and reducing reliance on dataset-specific context.}
Notably, VLCounter~\cite{DBLP:conf/aaai/KangMKH24} stands out among open-world text-guided methods (MAE: 6.46, RMSE: 8.68), but UPC’s performance remains notable given its unified prompt representation and refinement strategy.

\subsection{Textual Semantic Understanding}
\label{sec:sec:text_semantic_understanding}
Recently,~\cite{DBLP:journals/corr/abs-2409-15953} highlighted notable weaknesses in state-of-the-art open-world text-guided CAC methods. Specifically, the authors argued that these models often fail to accurately identify \textit{which} object class needs to be counted based on the textual description. Instead, they tend to count instances of the predominant class, disregarding the true intent of the prompt.
To address these shortcomings, the authors introduced a novel benchmark, including a test suite specifically designed to evaluate these limitations in current evaluation systems. 

\begin{table}[t]
    \centering
    \footnotesize
    \caption{\textbf{Textual semantic understanding results.} We illustrate the results on the FSC-147 dataset as reported in~\cite{DBLP:journals/corr/abs-2409-15953}, where the authors introduced a test suite designed to evaluate if open-world test-guided CAC methods are truly able to understand the provided textual prompts. The best results are in bold.}%
    \label{tab:text-semantic-understanding-results}%
    \newcolumntype{L}{>{\arraybackslash}m{.32\linewidth}}%
    \newcolumntype{C}{>{\centering\arraybackslash}X}
    %\tiny%
    \setlength{\tabcolsep}{2pt}
    \center
    \begin{tabularx}{0.95\linewidth}{L|CC|CC}
        \toprule
        & \multicolumn{2}{c|}{Validation Set} & \multicolumn{2}{c}{Test Set}\\
        \cmidrule(lr){2-3} \cmidrule(lr){4-5}
        \centering Method & NMN $\downarrow$ & PCCN $\uparrow$ & NMN $\downarrow$ & PCCN $\uparrow$ \\
        \midrule
        \midrule
        CounTX~\cite{AminiNaieni23} & 0.87 & 69.79 & 0.95 & 64.51 \\
        CLIP-Count~\cite{DBLP:conf/mm/JiangLC23} & 1.24 & 48.11 & 1.27 & 38.13 \\
        VLCounter~\cite{DBLP:conf/aaai/KangMKH24} & 1.07 & 62.70 & 1.15 & 53.36 \\
        TFPOC~\cite{DBLP:conf/wacv/Shi0Z24} & 0.67 & 62.62 & 0.75 & 66.04 \\
        DAVE~\cite{DBLP:conf/cvpr/PelhanLZK24} & \textbf{0.13} & \textbf{95.90} & \textbf{0.08} & \textbf{97.62} \\
        \bottomrule
    \end{tabularx}
\end{table}

Table~\ref{tab:text-semantic-understanding-results} presents results on the FSC-147 dataset, focusing on their negative-label test. This test evaluates the performance of the model on single-class images by prompting it with textual descriptions referencing absent object classes. \cite{DBLP:journals/corr/abs-2409-15953} introduced also two metrics for the assessment of this test: (i) the normalized mean of negative predictions (NMN), which is the absolute counting error computed by prompting the model with the negative classes, normalized by the ground truth of the positive class, and (ii) the positive class count nearness (PCCN) that mixes positive and negative class predictions, providing an overall quantitative assessment of strong failures in the model. More details can be found in~\cite{DBLP:journals/corr/abs-2409-15953}.
DAVE~\cite{DBLP:conf/cvpr/PelhanLZK24} performs exceptionally well across nearly all metrics proposed in~\cite{DBLP:journals/corr/abs-2409-15953}, standing out as the best-performing model in this challenging scenario. Although the authors noted that DAVE occasionally produces catastrophic errors due to sporadic outliers, it remains the top performer overall in this evaluation.

%\subsection{\textcolor{darkgreen}{Computational Complexity}}
\subsection{\textcolor{black}{Computational Complexity}}
\csname rev3\endcsname{
Table~\ref{tab:computational-complexity} reports the computational complexity---measured in GFLOPs and number of parameters---of the most representative CAC methods, i.e., those highlighted in Fig.~\ref{fig:temporal_line}. Specifically, we include values reported in the original papers when available, and compute them for methods that provide public code but omit complexity metrics. Models lacking both reported values and code are excluded from the analysis.}

\begin{table}[t]
    \centering
    \footnotesize
    \caption{\csname rev3\endcsname{\textbf{Computational complexity analysis.} We report the computational complexity of the most representative CAC methods in terms of GFLOPs and number of parameters. ``-'' indicates that the values were not computed because the authors did not provide a pretrained model. The best results are highlighted in bold.}}
    \label{tab:computational-complexity}%
    \newcolumntype{L}{>{\arraybackslash}m{.4\linewidth}}%
    \newcolumntype{C}{>{\centering\arraybackslash}X}
    %\tiny%
    \setlength{\tabcolsep}{2pt}
    \center
    \begin{tabularx}{0.95\linewidth}{L|C|C}
        \toprule
        \centering Method & GFLOPs $\downarrow$ & \#Parameters $\downarrow$ \\
        \midrule
        \midrule
        \textit{Reference-based} & & \\
        FamNet~\cite{DBLP:conf/cvpr/RanjanSNH21} & 55 & 26M \\
        RCAC$^{\ddagger}$~\cite{DBLP:conf/eccv/GongZ0DS22} & - & \textbf{10M} \\
        BMNet+~\cite{DBLP:conf/cvpr/Shi0FL022} & \textbf{27} & 13M \\
        SAFECount~\cite{DBLP:conf/wacv/YouYLLCL23} & 366 & 32M \\
        LOCA~\cite{DBLP:conf/iccv/EukicLZK23} & 80 & 37M \\
        SSD$^{\ddagger}$~\cite{DBLP:conf/ijcai/Xu0Z24} & - & 36M \\
        %CSTrans~\cite{DBLP:journals/pr/GaoH24} & - & 706k \\
        %CountVers~\cite{DBLP:journals/kbs/YangCDWZ24} & - & 35M \\
        CACViT~\cite{DBLP:conf/aaai/WangX0024} & 89 & 99M \\
        DAVE$^{\ddagger}$~\cite{DBLP:conf/cvpr/PelhanLZK24} & 1,515 & 47M \\
        TFPOC$^{\ddagger}$~\cite{DBLP:conf/wacv/Shi0Z24} & 102 & 94M \\
        CounTR$^{\ddagger}$$^{\$}$~\cite{DBLP:conf/bmvc/LiuZZX22} & 84 & 99M \\
        \midrule
        %\textit{Reference-less} & & \\
        %\midrule
        \textit{Open-world Text-guided} & & \\
        ZSC$^{\ddagger}$~\cite{DBLP:conf/cvpr/XuL0RS23} & - & 12M \\
        CounTX$^{\ddagger}$~\cite{AminiNaieni23} & 50 & 161M \\
        CLIP-Count$^{\ddagger}$~\cite{DBLP:conf/mm/JiangLC23} & \textbf{27} & 166M \\
        VLCounter~\cite{DBLP:conf/aaai/KangMKH24} & 68 & \textbf{1.4M} \\
        GroundingREC$^{\ddagger}$~\cite{DBLP:conf/cvpr/DaiLC24} & 69 & 173M \\
        VA-Count$^{\ddagger}$~\cite{DBLP:conf/eccv/ZhuYYGWZH24} & - & 99M \\
        %DAVE~\cite{DBLP:conf/cvpr/PelhanLZK24} & TODO & TODO \\
        \bottomrule
    \end{tabularx}
    \par\vspace{4pt}
    {\setlength{\baselineskip}{0.7\baselineskip}\footnotesize
    $\ddagger$ computed in this work; \$ reference-based modification of a reference-less method.}
\end{table}

\csname rev3\endcsname{The results reveal substantial variability in computational requirements across paradigms and architectures.
Among reference-based methods, models such as FamNet~\cite{DBLP:conf/cvpr/RanjanSNH21} and BMNet+~\cite{DBLP:conf/cvpr/Shi0FL022} are relatively lightweight, requiring 55 and 27 GFLOPs, respectively, with moderate parameter counts (26M and 13M). In contrast, more recent architectures like SAFECount~\cite{DBLP:conf/wacv/YouYLLCL23} and LOCA~\cite{DBLP:conf/iccv/EukicLZK23} show a marked increase in both GFLOPs (366 and 80) and parameters (32M and 37M), reflecting the adoption of deeper backbones and more sophisticated matching modules. DAVE~\cite{DBLP:conf/cvpr/PelhanLZK24} represents a particularly interesting case since it stands out as the state-of-the-art in terms of accuracy, achieving the lowest MAE and RMSE on both FSC-147~\cite{DBLP:conf/cvpr/RanjanSNH21} and CARPK~\cite{DBLP:conf/iccv/HsiehLH17}. However, this performance comes at a very high computational cost: 1515 GFLOPs and 47M parameters---an order of magnitude higher than most other methods. This makes DAVE ideal for high-resource environments, but potentially unsuitable for edge deployment.}

\csname rev3\endcsname{In the open-world text-guided paradigm, computational demands are generally higher due to the integration of large vision-language models. For instance, CounTX~\cite{AminiNaieni23} and GroundingREC~\cite{DBLP:conf/cvpr/DaiLC24} reach parameter counts of 161M and 173M, respectively. Notably, VLCounter~\cite{DBLP:conf/aaai/KangMKH24} achieves a remarkably low parameter count (1.4M) while maintaining competitive GFLOPs (68), thanks to efficient prompt-tuning and segment-aware skip connections.}

\csname rev3\endcsname{Overall, the analysis reveals that higher computational complexity does not always correlate with better performance. While DAVE justifies its cost with top-tier accuracy, other models like CACViT, SSD, and BMNet+ offer a more favorable trade-off between efficiency and accuracy. These findings suggest that architectural innovations---rather than brute-force scaling---are key to achieving practical and scalable CAC solutions.}

\begin{figure*}[!t]
    \centering
    \includegraphics[width=.95\linewidth]{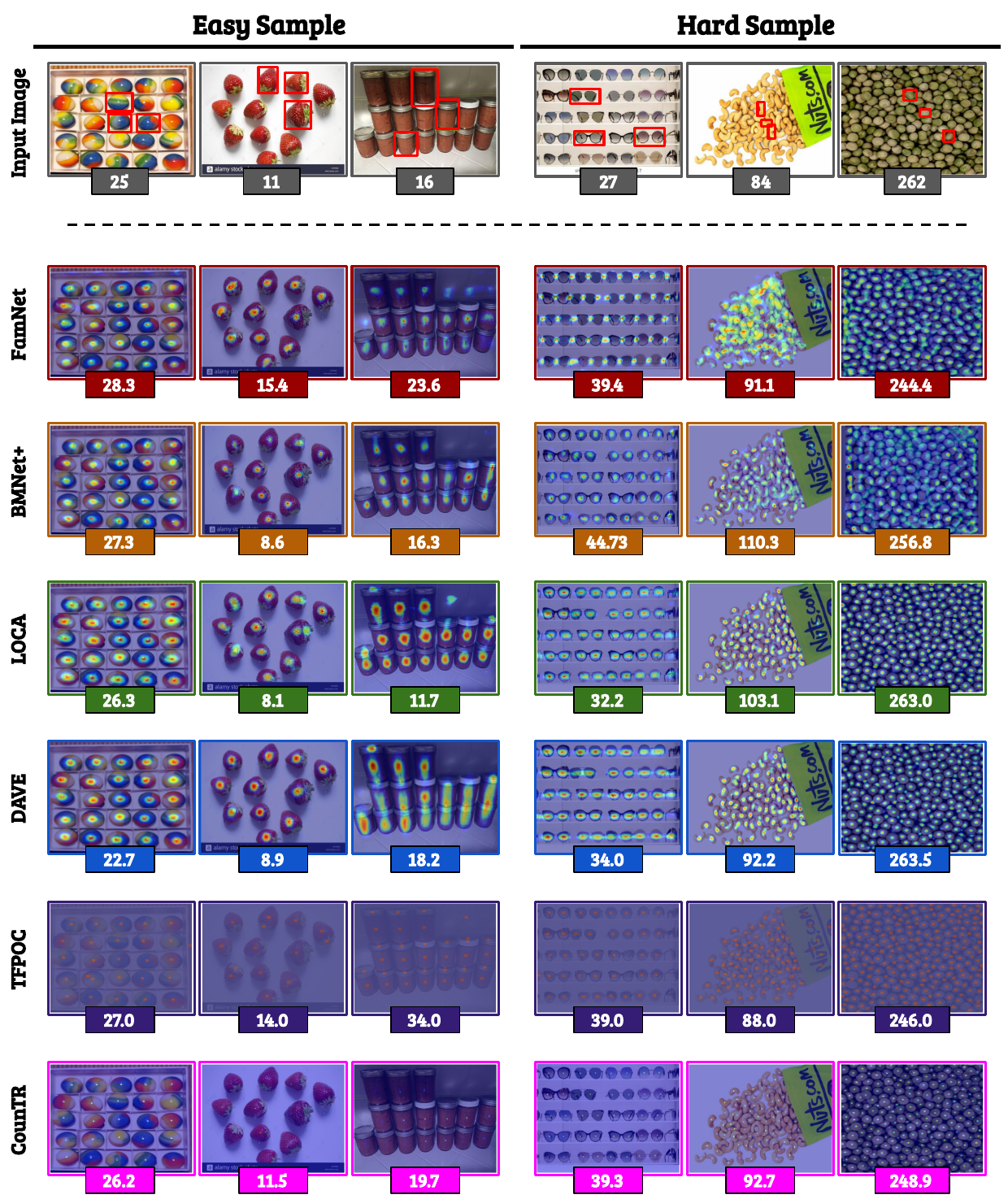}
    \caption{\csname rev3\endcsname{\textbf{Qualitative results of reference-based CAC methods.} We report some qualitative results of the most representative reference-based CAC techniques. On the left, we show three easy samples, while on the right, three challenging samples that result in large counting errors.}}
    \label{fig:qualitative-results-few-shot}
\end{figure*}

\begin{figure*}[!t]
    \centering
    \includegraphics[width=.95\linewidth]{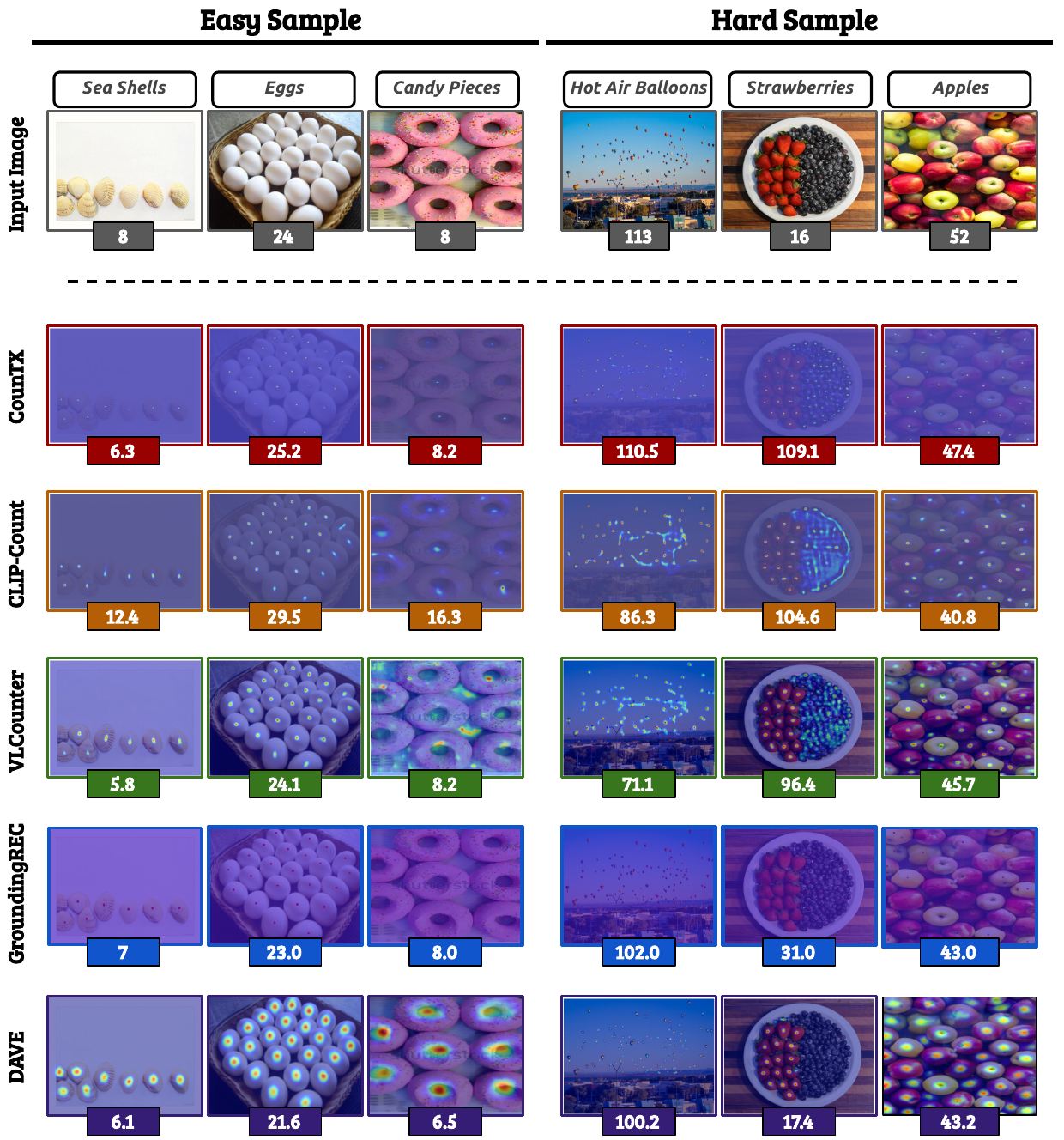}
    \caption{\csname rev3\endcsname{\textbf{Qualitative results of open-world text-guided CAC methods.} We report some qualitative results of the most representative open-world text-guided CAC techniques. On the left are three easy samples, while on the right are three challenging samples that result in large counting errors.}}
    \label{fig:qualitative-results-text}
\end{figure*}

%\subsection{\textcolor{darkgreen}{Qualitative Comparison}}
\subsection{\textcolor{black}{Qualitative Comparison}}
\csname rev3\endcsname{Fig.~\ref{fig:qualitative-results-few-shot} presents qualitative results for reference-based CAC techniques. We compare the outputs of the most representative CAC models (see Fig.~\ref{fig:temporal_line}) for which pretrained weights are publicly available. Specifically, we show both easy and hard samples, with the latter often corresponding to failure cases.
Note that, for TFPOC~\cite{DBLP:conf/wacv/Shi0Z24}, density maps were generated from the predicted masks by placing a small circle at the centroid of each mask.} 

\csname rev3\endcsname{Overall, DAVE~\cite{DBLP:conf/cvpr/PelhanLZK24} confirms its state-of-the-art performance. However, it is worth noting that even DAVE, like all other models, struggles with certain challenging samples.
One recurring challenge appears to be related to object density---objects tend to overlap in crowded scenes. This issue becomes more pronounced when additional elements such as background clutter and distractors are present, as in the second hard sample. In contrast, it is less evident in scenarios where only the target objects are present and are relatively uniformly distributed, as in the third hard sample.
A particularly interesting failure case is illustrated by the first hard sample. Despite the absence of occlusion and the seemingly simple scene, all models consistently fail by counting the lenses of the glasses rather than the glasses as whole objects. This suggests that the models are unable to build effective representations of the exemplars and instead rely on detecting repetitive patterns in the image.}

\csname rev3\endcsname{Fig.~\ref{fig:qualitative-results-text} presents instead qualitative results for the most representative open-world text-guided CAC methods. Note that, for GroundingREC~\cite{DBLP:conf/cvpr/DaiLC24}, density maps were generated by placing a small circle at the centroids of the generated bounding boxes. The most pronounced and interesting failure case is observed in the second hard sample. As previously noted by~\cite{DBLP:journals/corr/abs-2409-15953} and discussed in Sec.~\ref{sec:sec:text_semantic_understanding}, this error stems from the limited ability of the models to understand the semantic content of textual descriptions. In particular, they tend to count repeated objects regardless of the actual meaning conveyed by the text. This issue is clearly illustrated in the image, which contains a large number of objects from two distinct categories. Only DAVE~\cite{DBLP:conf/cvpr/PelhanLZK24}, and to a lesser extent GroundingREC~\cite{DBLP:conf/cvpr/DaiLC24}, can successfully address this challenge.}

\section{Conclusion and Open Challenges}
\label{sec:conclusion}
Class-agnostic counting has become a pivotal field in object counting, addressing the limitations of traditional class-specific methods. Unlike conventional approaches that require extensive labeled data for each object category, CAC enables models to count objects across arbitrary categories, including those unseen during training. For the first time, this survey reviewed the evolution of CAC methodologies, categorizing them into three paradigms: reference-based, reference-less, and open-world text-guided approaches. Each paradigm brings distinct advantages and trade-offs, reflecting the diverse demands of real-world applications.
\csname rev1\endcsname{Reference-based methods, which rely on annotated exemplars, set the benchmark for accuracy and robustness. However, their dependence on labeled exemplars limits scalability and automation. Reference-less approaches address this by automatically identifying dominant object classes, offering greater flexibility but still lagging behind in performance. Open-world text-guided methods further increase adaptability by allowing object specification via textual prompts, yet---like reference-less methods---this flexibility comes at the cost of reduced accuracy compared to exemplar-based approaches.}

\csname rev1\endcsname{Through an in-depth analysis of state-of-the-art methods on benchmarks such as FSC-147~\cite{DBLP:conf/cvpr/RanjanSNH21} and CARPK~\cite{DBLP:conf/iccv/HsiehLH17}, this survey highlights both notable advancements and persistent challenges. While DAVE~\cite{DBLP:conf/cvpr/PelhanLZK24} leads the reference-based paradigm on FSC-147 (MAE: 8.66, RMSE: 32.36), the reference-less and text-guided leaderboards remain less conclusive. CounTR~\cite{DBLP:conf/bmvc/LiuZZX22} and GCNet~\cite{DBLP:journals/pr/WangLZTG24} stand out among reference-less methods, the latter notable for its weak supervision. In the text-guided setting, GroundingREC~\cite{DBLP:conf/cvpr/DaiLC24} achieves the lowest MAE, while DAVE remains robust to outliers.
Cross-dataset evaluations on CARPK reveal that CACViT~\cite{DBLP:conf/aaai/WangX0024}, UPC~\cite{DBLP:conf/aaai/0018C24}, and VLCounter~\cite{DBLP:conf/aaai/KangMKH24} generalize best across paradigms. Qualitative results further confirm that even top-performing models can fail in cluttered or ambiguous scenes, often due to poor or missing guidance. Moreover, recent benchmarks~\cite{DBLP:journals/corr/abs-2409-15953} show that open-world models frequently misinterpret prompts, highlighting limitations in semantic understanding.}

\csname rev1\endcsname{Open challenges span all paradigms. A major limitation is the lack of diverse and publicly available datasets.} The gold standard, FSC-147, has several drawbacks. First, it contains a limited number of images and object classes. Furthermore, as pointed out in~\cite{DBLP:journals/corr/abs-2409-15953}, almost all images contain only a single class of objects. Multi-class images would be a desirable feature for new datasets, as objects belonging to classes different from the one being counted could act as distractors for the counting approach. 
The lack of diverse and large-scale datasets also leads to challenges in designing learning approaches suited for data-scarce scenarios. Amnog the analyzed methods, only RCC~\cite{DBLP:journals/corr/abs-2205-10203} and GCNet~\cite{DBLP:journals/pr/WangLZTG24} operate with weak supervision, avoiding the use of dot annotations. This setting is particularly promising, as it could help mitigate the challenge of creating large-scale labeled datasets for the counting task. Recently, an emerging direction is to incorporate completely unsupervised components into CAC pipelines~\cite{DBLP:journals/corr/abs-2504-16570,DBLP:conf/cvpr/KnobelHA24}, although current performance remains far from state-of-the-art.

\csname rev3round2\endcsname{Other challenges are more specific to open-world text-guided methodologies, which appear to represent the future. These models have been shown to occasionally misinterpret the meaning of prompts. Although some architectures demonstrate relative robustness, further improvements are needed. New architectural designs and learning strategies---such as contrastive learning over classes not present in the images---should be explored. Another particularly interesting direction concerns the use of richer textual inputs to enable context-aware prompting. Future research could investigate automatic prompt generation techniques to produce more informative and targeted textual queries. While current large multimodal models are not yet effective for counting tasks, partly due to limitations in visual token capacity, their contextual priors and exposure to diverse visual-textual data suggest potential for future exploration, especially as architectures evolve to better handle fine-grained visual details.} \csname rev3\endcsname{Beyond these technical challenges, our analysis underscores a broader issue: the convenience offered by reference-less and text-guided paradigms comes at a measurable cost in accuracy. Bridging this gap---by developing methods that match the guidance quality of human-provided exemplars without manual intervention---remains a key open challenge.
Finally, our computational complexity analysis reveals that top-performing models such as DAVE achieve state-of-the-art accuracy at the cost of significantly higher GFLOPs and parameter counts. This highlights a critical trade-off between performance and efficiency, which must be carefully considered when designing CAC systems for real-world deployment, particularly in resource-constrained environments.}

%Across paradigms, a clear ranking of models based on performance metrics does not always occur.
%DAVE leads the field in reference-based counting, achieving the lowest MAE (8.66) and RMSE (32.36) on FSC-147. Its detect-and-verify strategy sets a benchmark for accuracy and robustness. Among reference-less models, CounTR demonstrates competitive performance with an MAE of 14.12 and RMSE of 108.01, showcasing the effectiveness of its transformer-based design. For open-world text-specified counting, CounTX stands out with the best MAE (15.73) and RMSE (106.88), highlighting the promise of vision-language integration. These results underline the strengths and weaknesses of each paradigm. Reference-based methods dominate in precision, but their reliance on annotated exemplars restricts scalability. Reference-less models offer greater flexibility but require further refinement to close the accuracy gap. Open-world text-specified approaches show significant potential, especially in applications demanding adaptability to diverse object categories.

\bibliographystyle{IEEEtran}
\bibliography{biblio}

%Biography text here without a photo.

%\begin{IEEEbiography}[{\includegraphics[width=1in,height=1.25in,clip,keepaspectratio]{fig1.png}}]{IEEE Publications Technology Team}
%In this paragraph you can place your educational, professional background and research and other interests.\end{IEEEbiography}

\end{document}